\pdfoutput=1
\documentclass[10pt,logo,copyright]{nvidiatechreport}
\usepackage{algorithm}

\usepackage{parskip}        %
\usepackage{amsfonts}       %
\usepackage{nicefrac}       %
\usepackage{animate}        %
\usepackage{subcaption}
\usepackage{tabularx}
\usepackage{makecell}
\usepackage{adjustbox}
\usepackage{setspace}
\newcolumntype{M}[1]{>{\centering\arraybackslash}m{#1}}
\usepackage{float}
\usepackage{tikz}
\usetikzlibrary{positioning,shapes,arrows}
\usepackage{amsmath,amsfonts,bm, bbm,leftindex}
\usepackage{multirow}
\usepackage{comment}
\usepackage{gensymb}
\usepackage{lipsum}
\usetikzlibrary{arrows.meta, positioning, fit}
\usepackage[para]{threeparttable}
\usepackage{tikz}
\usetikzlibrary{tikzmark}

\let\cite\citep
\usepackage{natbib}

\definecolor{darkred}{rgb}{0.7, 0.0, 0.0}

\usepackage{pifont}
\usepackage{wrapfig}

\usepackage[nameinlink]{cleveref}
\crefname{equation}{Eq.}{Eqs.}
\crefname{figure}{Fig.}{Figs.}
\crefname{section}{Sec.}{Sec.}
\crefname{appendix}{App.}{App.}
\crefname{table}{Tab.}{Tabs.}
\crefname{algorithm}{Algo}{Algo}
\crefname{thm}{Thm}{Thm}
\Crefname{thm}{Thm}{Thm}
\crefname{prop}{Prop}{Prop}
\usepackage{longtable}
\usepackage{mathtools}
\usepackage{siunitx}

\usepackage{ragged2e}
\definecolor{nvidiagreen}{HTML}{76B900}
\definecolor{bestrow}{HTML}{E1EBD7}
\newcolumntype{Y}{>{\raggedleft\arraybackslash}X}
\newcolumntype{L}[1]{>{\raggedright\arraybackslash}p{#1}}

\usepackage{tikz}
\usepackage{siunitx}
\usepackage{algorithm}
\usepackage{algpseudocode}
\usepackage{comment}
\definecolor{nvidiaGreen}{RGB}{118,185,0}
\usepackage{appendix}
\usepackage{minitoc}

\usepackage{titletoc}
\usepackage{wrapfig}
\Crefname{figure}{Fig.}{Figs.}
\Crefname{table}{Tab.}{Tab.}

\usetikzlibrary{positioning, calc, shapes, arrows.meta, backgrounds, spy}
\dosecttoc

\newcommand{\methodname}{\textbf{\textcolor{nvidiaGreen}{Motive}}\xspace}

\usepackage{misc/annotate_equations}

\usepackage{amsmath,amsfonts,bm}

\def\eqref#1{Eq.~\ref{#1}}

\def\1{\bm{1}}

\DeclareMathAlphabet{\mathsfit}{\encodingdefault}{\sfdefault}{m}{sl}
\SetMathAlphabet{\mathsfit}{bold}{\encodingdefault}{\sfdefault}{bx}{n}

\newcommand{\E}{\mathbb{E}}

\newcommand{\numFrames}{F}                 %
\newcommand{\frameIndex}{f}                %
\newcommand{\videoHeight}{H}               %
\newcommand{\videoHeightIndex}{h}          %
\newcommand{\videoWidth}{W}                %
\newcommand{\videoWidthIndex}{w}           %
\newcommand{\timestep}{t}                  %
\newcommand{\timestepSize}{T}              %
\newcommand{\identity}{\mathbf{I}}         %
\newcommand{\videoFrame}{\mathbf{f}}       %

\newcommand{\dataset}{\mathcal{D}}         %
\newcommand{\datasetft}{\dataset_{\textnormal{ft}}}        %
\newcommand{\datasetSize}{N}               %
\newcommand{\datasetIndex}{n}              %
\newcommand{\dataSubset}{\mathcal{S}}      %
\newcommand{\queryVideoclip}{\hat{\videoclip}}     %
\newcommand{\queryConditioning}{\hat{\conditioning}} %
\newcommand{\influenceIndex}{k}            %
\newcommand{\influenceSubsetSize}{K}       %
\newcommand{\numericalBias}{\zeta}         %
\newcommand{\numMotions}{Q} %
\newcommand{\motionIndex}{q} %

\newcommand{\modelparams}{\boldsymbol{\theta}}     %
\newcommand{\videogenerator}{p_{\modelparams}}      %
\newcommand{\videoclip}{\mathbf{v}}                %
\newcommand{\conditioning}{\mathbf{c}}             %
\newcommand{\vaeencoder}{E}                        %
\newcommand{\latentChannelSize}{C}                 %
\newcommand{\latentDownsampleFactor}{s}            %
\newcommand{\latents}{\mathbf{h}}                  %
\newcommand{\latentHeightIndex}{\tilde{h}}         %
\newcommand{\latentWidthIndex}{\tilde{w}}          %

\newcommand{\noisevector}{\boldsymbol{\epsilon}}               %
\newcommand{\noisepredict}{\noisevector_{\modelparams}}        %
\newcommand{\timestepfixed}{\timestep_{\textnormal{fix}}}      %
\newcommand{\fixednoise}{\noisevector_{\textnormal{fix}}}      %
\newcommand{\noisyLatents}{\mathbf{z}}                         %
\newcommand{\vectorfield}{\mathbf{f}_{\modelparams}}           %
\newcommand{\timederivative}{\dot{\noisyLatents}}              %
\newcommand{\schedulerCoeffA}{\alpha_{\timestep}}              %
\newcommand{\schedulerCoeffB}{\sigma_{\timestep}}              %
\newcommand{\trainingtarget}{\noisevector_{\textnormal{target}}} %

\newcommand{\loss}{\mathcal{L}}                    %
\newcommand{\lossdiff}{\loss_{\textnormal{diff}}}  %
\newcommand{\lossflow}{\loss_{\textnormal{flow}}}  %
\newcommand{\lossmotionweighted}{\loss_{\textnormal{mot}}} %
\newcommand{\perLocError}{\tilde{\loss}}           %

\newcommand{\gradient}{\mathbf{g}}                 %
\newcommand{\gradientProjected}{\tilde{\gradient}} %
\newcommand{\gradientMotion}{\gradient_{\textnormal{mot}}}                  %
\newcommand{\gradientMotionProjected}{\gradientProjected_{\textnormal{mot}}} %
\newcommand{\influenceFunc}{I}                     %
\newcommand{\InfluenceMot}{\influenceFunc_{\textnormal{mot}}} %
\newcommand{\hessianmatrix}{\mathbf{H}_{\modelparams}}        %
\newcommand{\spearmancorr}{\rho}                   %
\newcommand{\TopK}{\operatorname{TopK}}            %
\newcommand{\MajVote}{\operatorname{MajVote}}      %
\newcommand{\influenceThreshold}{\tau}             %

\newcommand{\gradientdim}{D}                       %
\newcommand{\projecteddim}{\gradientdim'}          %
\newcommand{\projectionMatrix}{\mathbf{P}}         %
\newcommand{\fastfoodNorm}{\xi}                    %
\newcommand{\timestepset}{\mathcal{T}}             %
\newcommand{\cost}{B}                              %

\newcommand{\motionFlow}{\mathbf{F}}               %
\newcommand{\displacementvec}{\mathbf{D}_{\!\frameIndex}}     %
\newcommand{\motionmagnitude}{M_{\!\frameIndex}}              %
\newcommand{\motionweights}{\mathbf{W}}            %
\newcommand{\alltracker}{\mathcal{A}}              %
\newcommand{\motioninfo}{A}                        %
\newcommand{\motionspace}{\mathbb{R}^{\numFrames \times \videoHeight \times \videoWidth \times 4}} %

\newcommand{\inputData}{\mathbf{x}}                %
\newcommand{\testdata}{\inputData_{\textnormal{test}}}   %
\newcommand{\trainndata}{\inputData_{\datasetIndex}}     %
\newcommand{\trainVideoData}{\videoclip_{\datasetIndex}} %

\usepackage{tcolorbox}
\definecolor{chart}{HTML}{1f77b4}

\definecolor{nvidiagreen}{RGB}{118, 185, 0}
\newtcolorbox{example}[1][]{
  colback=white,
  colframe=nvidiagreen,
  floatplacement=floating,
  fonttitle=\bfseries,
  title=\centering \textsf{#1}
}
\definecolor{amethyst}{rgb}{0.6, 0.4, 0.8}

\definecolor{lemon}{RGB}{255,247,0}
\definecolor{maize}{RGB}{250,237,94}
\definecolor{coco1}{HTML}{D9E4EC}
\definecolor{coco2}{HTML}{B7CFDC}
\definecolor{coco3}{HTML}{6AABD2}
\definecolor{coco4}{HTML}{385E72}

\definecolor{mustard}{RGB}{255,219,89}
\definecolor{ocre}{RGB}{241,103,35}
\definecolor{Tangerine}{RGB}{253,128,8}
\definecolor{framegreen}{RGB}{153, 188, 133}
\definecolor{bggreen}{RGB}{235, 250, 228}
\definecolor{lightgray}{RGB}{239,240,241}
\definecolor{c0}{cmyk}{1,0.3968,0,0.2588} 
\definecolor{c1}{cmyk}{0,0.6175,0.8848,0.1490} 
\definecolor{c2}{cmyk}{0.1127,0.6690,0,0.4431} 
\definecolor{c3}{cmyk}{0.3081,0,0.7209,0.3255} 
\definecolor{c4}{RGB}{164, 16, 52}

\definecolor{g0}{HTML}{ffdbdc}
\definecolor{g1}{HTML}{ffefeb}
\definecolor{g2}{HTML}{FCFFEB}
\definecolor{g3}{HTML}{e9fcea}
\definecolor{g4}{HTML}{d4f8d4}

\definecolor{c0}{cmyk}{1,0.3968,0,0.2588} 
\definecolor{c1}{cmyk}{0,0.6175,0.8848,0.1490} 
\definecolor{c2}{cmyk}{0.1127,0.6690,0,0.4431} 
\definecolor{c3}{cmyk}{0.3081,0,0.7209,0.3255} 
\definecolor{c4}{RGB}{164, 16, 52}
\definecolor{orange}{HTML}{E66100}
\definecolor{bluex}{HTML}{0C7BDC}
\definecolor{yellow}{HTML}{FFC20A}
\definecolor{lightpurple}{HTML}{E6E6FA}
\definecolor{lightbluee}{HTML}{e8f4f8}
\definecolor{c5}{HTML}{EE4E4E}
\definecolor{gggggg}{HTML}{EFEFEF}
\newtcbox{\hlprimarytab}{on line, box align=base, colback=blue!12,colframe=white,size=fbox,arc=3pt, before upper=\strut, top=-2pt, bottom=-4pt, left=-2pt, right=-2pt, boxrule=0pt}
\newtcbox{\hlsecondarytab}{on line, box align=base, colback=orange!10,colframe=orange,size=fbox,arc=3pt, before upper=\strut, top=-2pt, bottom=-4pt, left=-2pt, right=-2pt, boxrule=0pt}
\newtcbox{\hlwhite}{on line, box align=base, colback=blue!12,colframe=white,size=fbox,arc=2pt, before upper=\strut, top=-2pt, bottom=-4pt, left=-2pt, right=-2pt, boxrule=0pt}
\newtcbox{\hlyellow}{on line, box align=base, colback=BlueGreen!10,colframe=white,size=fbox,arc=2pt, before upper=\strut, top=-2pt, bottom=-4pt, left=-2pt, right=-2pt, boxrule=0pt}

\usepackage[framemethod=TikZ]{mdframed}
\usepackage{tcolorbox}
\definecolor{chart}{HTML}{1f77b4}
\mdfsetup{%
    middlelinecolor =   none,
    middlelinewidth =   1pt,
    backgroundcolor =   blue!5,
    roundcorner     =   5pt,
}
\usepackage{colortbl}
\newcommand{\sbt}[1]{\textbf{\texttt{\small #1}}}
\usepackage{footnote}
\makesavenoteenv{table}

\usepackage{silence}
\usepackage{dsfont}
\usepackage{mathtools}
\usepackage{subcaption}
\usepackage{placeins}
\usepackage{stackengine}
\usepackage{multirow}
\usepackage{adjustbox}
\usepackage[framemethod=TikZ]{mdframed}	
\usepackage{setspace}  
\usepackage{multicol} 
\usepackage{transparent}

\definecolor{gblue}{RGB}{66,133,244}
\definecolor{gred}{RGB}{219,68,55}
\definecolor{gyellow}{RGB}{244,180,0}
\definecolor{ggreen}{RGB}{15,157,88}
\definecolor{lpcolor}{RGB}{42,74,138}
\definecolor{morelcolor}{RGB}{185,18,32}
\definecolor{bgcolor}{RGB}{230,245,208}
\definecolor{framecolor}{RGB}{244,109,67}
\definecolor{mulberry}{rgb}{0.77, 0.29, 0.55}
\definecolor{IndianRed2}{RGB}{238,121,121}
\definecolor{Green4}{RGB}{0,139,0}
\colorlet{cred}{IndianRed2}
\colorlet{cgreen}{Green4}

\usepackage{arydshln}
\makeatletter
\def\adl@drawiv#1#2#3{%
        \hskip.5\tabcolsep
        \xleaders#3{#2.5\@tempdimb #1{1}#2.5\@tempdimb}%
                #2\z@ plus1fil minus1fil\relax
        \hskip.5\tabcolsep}
\newcommand{\cdashlinelr}[1]{%
  \noalign{\vskip\aboverulesep
           \global\let\@dashdrawstore\adl@draw
           \global\let\adl@draw\adl@drawiv}
  \cdashline{#1}
  \noalign{\global\let\adl@draw\@dashdrawstore
           \vskip\belowrulesep}}
\makeatother

\usepackage{tcolorbox}
\usetikzlibrary{shadows}

\tcbset{
    myquote/.style={
        enhanced,
        colback=white,    %
        colframe=black,    %
        leftrule=1mm,     %
        toprule=-0.1mm,      %
        bottomrule=-0.1mm,   %
        rightrule=-0.1mm,    %
        sharp corners,    %
        left=1.2mm          %
    }
}

\makeatletter
\newcommand\ttiny{\@setfontsize\ttiny{5}{5}}
\makeatother

\newcommand{\hide}[1]{}

\makeatletter
\graphicspath{{fig/}}
\makeatother

\makeatletter
\newcommand{\crefnames}[3]{%
  \@for\next:=#1\do{%
    \expandafter\crefname\expandafter{\next}{#2}{#3}%
  }%
}
\makeatother

\title{Motion Attribution for Video Generation}

\begin{document}

\author{
  Xindi Wu$^{1,2}$,
  Despoina Paschalidou$^{1}$,
  Jun Gao$^{1,4}$,
  Antonio Torralba$^{3}$,
  Laura Leal-Taixé$^{1}$,
  Olga Russakovsky$^{2}$,
  Sanja Fidler$^{1,5,6}$,
  Jonathan Lorraine$^{1}$\\
  \small $^{1}$NVIDIA \quad
         $^{2}$Princeton University \quad
         $^{3}$MIT \quad
         $^{4}$University of Michigan \quad
         $^{5}$University of Toronto \quad 
         $^{6}$Vector Institute\\
  \small \href{https://research.nvidia.com/labs/sil/projects/MOTIVE/}{\textcolor{nvidiaGreen}{https://research.nvidia.com/labs/sil/projects/MOTIVE/}}
}

\maketitle
\vspace{-0.01\textheight}
\begin{abstract}
Despite the rapid progress of video generation models, the role of data in influencing motion is poorly understood. We present \methodname (\textbf{\underline{MOTI}on attribution for \underline{V}ideo g\underline{E}neration}), a motion-centric, gradient-based data attribution framework that scales to modern, large, high-quality video datasets and models.
We use this to study which fine-tuning clips improve or degrade temporal dynamics.
\methodname\ isolates temporal dynamics from static appearance via motion-weighted loss masks, yielding efficient and scalable motion-specific influence computation.
On text-to-video models, \methodname\ identifies clips that strongly affect motion and guides data curation that improves temporal consistency and physical plausibility. 
With \methodname-selected high-influence data, we improve both motion smoothness and dynamic degree on VBench, achieving a 74.1\% human preference win rate compared with the pretrained base model.
To our knowledge, this is the first framework to attribute motion rather than visual appearance in video generative models and to use it to curate fine-tuning data.
\end{abstract}

\abscontent
\section{Introduction}

Motion is the defining element of videos. Unlike image generation, which produces a single frame, video generative models capture how objects move, interact, and obey physical constraints~\citep{wiedemer2025video, kang2024far}. Yet even with the rapid progress of video generation, a fundamental question remains: \\
\vspace{-5pt}
\begin{mdframed}[leftmargin=0pt, rightmargin=0pt, innerleftmargin=10pt, innerrightmargin=10pt, innertopmargin=6pt, innerbottommargin=6pt, skipbelow=0pt]
\begin{center}
\emph{Which training data influence the motion in generated videos?}
\end{center}
\end{mdframed}
\vspace{-5pt}
\textbf{Why it matters.}
Diffusion models are data-driven, and their progress has tracked the scaling of data and compute~\citep{saharia2022photorealistic, nichol2021improved, ho2022video, peebles2023scalable}. 
Prior work~\citep{blattmann2023stable, kaplan2020scaling, ravishankar2025scaling} shows that training data shapes key generative properties, including visual quality~\citep{rombach2022high}, semantic fidelity~\citep{namekata2024emerdiff}, and compositionality~\citep{wu2025compact, favero2025compositional}.
Motion is no exception. \textit{Motion} refers to temporal dynamics captured by optical flow, including trajectories, deformations, camera movement, and interactions. If generated motion reflects the data distribution that shaped the model, then attributing motion to influential training clips provides a direct lens on why a model moves the way it does and enables targeted data selection for desired dynamics.
High-quality data often matters most in fine-tuning, where large pretraining corpora are inaccessible and carefully selected clips can have an outsized impact. Motion-specific attribution is therefore especially valuable in the fine-tuning regime, where the goal is to identify which clips most influence temporal coherence and physical plausibility.

\textbf{Why existing methods are limited for motion.}
Prior diffusion data attribution focuses on images and explains static content. Extending these methods to videos na\"ively collapses motion into appearance, missing the temporal structure that distinguishes videos from images. Three challenges drive this gap: (i) localizing motion so attribution focuses on dynamic regions rather than static backgrounds, (ii) scaling to sequences since gradients must integrate across time, and (iii) capturing temporal relations like velocity, acceleration, and trajectory coherence that single-frame attribution cannot measure. Addressing motion attribution requires methods that explicitly model temporal structure, rather than treating time as an additional spatial axis.

\textbf{Our method.}
We introduce \methodname, a motion attribution framework for video generation models that isolates motion-specific influence. \methodname computes gradients with motion-aware masking. As a result, the attribution signal emphasizes dynamic regions rather than static appearance. Efficient approximations make the method practical for large, high-quality datasets and video generative models. Our scores trace generated motion back to training clips, enabling targeted curation and improving motion quality when used to guide fine-tuning. Our key contributions are:

\begin{enumerate}
\setlength{\itemsep}{0pt}
\setlength{\parskip}{0pt}
\vspace{-0.000\textheight}
\item Proposing a scalable gradient-based attribution approach for video generation models that is computationally efficient, even at the scale of modern, high-quality data and large generative models (\S\ref{sec:method:1}).
\item Addressing a video-specific bias by correcting frame-length effects in gradient magnitudes, ensuring fair attribution across clips of different durations (\S\ref{sec:method:video-bias}).
\item Introducing an attribution that emphasizes temporal dynamics to trace which training clips most strongly influence motion quality (\S\ref{sec:method:2}). 
\item Showing that we improve motion smoothness and dynamic degree on VBench and in human evaluation  (\S\ref{sec:experiment}), matching, or surpassing, full-dataset fine-tuning performance with only $10\%$ of the data, and outperforming motion-unaware attribution baselines (Tables~\ref{tab:vbench-wan} and \ref{tab:human_eval_extended}).
\end{enumerate}

\section{Background}
\label{sec:background}

Notation and extended related work are in App.~\S\ref{app:notation} and \S\ref{sec:app:related}.
\subsection{Video Generation with Diffusion and Flow-Matching Models}
\label{subsec:diffusion}

\textbf{Diffusion and flow matching in latent space.}
Let $\videogenerator(\videoclip \mid \conditioning)$ be a conditional generator with parameters $\modelparams$, where $\videoclip \in \mathbb{R}^{\numFrames \times \videoHeight \times \videoWidth \times 3}$ is a clip of height $\videoHeight$, width $\videoWidth$, and $\numFrames$ frames, and $\conditioning$ denotes conditioning such as text or other multimodal metadata (e.g., fps, depth, pose).
We operate in VAE latents: $\latents=\vaeencoder(\videoclip)$ and train a denoiser or velocity field on noisy latents.
A noise scheduler supplies time-dependent coefficients $(\schedulerCoeffA,\schedulerCoeffB)$ controlling signal and noise scales, and the forward noising is:\vspace{-0.005\textheight}
\begin{equation}
\smash{
\noisyLatents(\timestep, \noisevector) \;=\; \schedulerCoeffA\,\latents \;+\; \schedulerCoeffB\,\noisevector,
\quad \noisevector \sim \mathcal{N}(0,\identity), \quad \timestep \in \{1,\ldots,\timestepSize\}.
}
\label{eq:forward}\vspace{-0.005\textheight}
\end{equation}
\emph{Denoising diffusion}~\citep{ho2020denoising} trains a network $\noisepredict(\noisyLatents,\conditioning,\timestep)$ to predict the injected noise:\vspace{-0.005\textheight}
\begin{equation}
\smash{
\lossdiff(\modelparams;\videoclip,\conditioning)
\;=\;
\E_{\timestep,\noisevector}\!\left[
\left\lVert
\noisepredict(\noisyLatents(\timestep, \noisevector),\conditioning,\timestep) - \noisevector
\right\rVert_2^2
\right].
}
\label{eq:ldiff}\vspace{-0.005\textheight}
\end{equation}
\emph{Flow matching}~\citep{lipman2022flow,albergo2023stochastic} learns a time-dependent vector field $\vectorfield(\noisyLatents_\timestep,\conditioning,\timestep)$ that matches the instantaneous velocity $\timederivative=\tfrac{\textnormal{d}}{\textnormal{d}\timestep}\noisyLatents$ induced by a chosen interpolant:\vspace{-0.005\textheight}
\begin{equation}
\smash{
\lossflow(\modelparams;\videoclip,\conditioning)
\;=\;
\E_{\timestep,\noisevector}\!\left[
\left\lVert
\vectorfield(\noisyLatents(\timestep, \noisevector),\conditioning,\timestep) - \timederivative(\timestep, \noisevector)
\right\rVert_2^2
\right].
}
\label{eq:lflow}\vspace{-0.005\textheight}
\end{equation}
Both objectives train time-indexed predictors over the latent space by integrating over $\timestep$ and $\noisevector$, thus gradient-based methods like attribution share similar challenges.

\textbf{From images to video for generation.}
Adding a temporal axis materially changes modeling and training.
Generation must capture spatial appearance and temporal dynamics such as object and camera motion, deformations, and interactions.
Modern systems extend image backbones with temporal capacity, for example, 3D U-Nets or 2D U-Nets augmented with temporal attention, causal or sliding-window context, and factorized space-time blocks, often trained in a latent-video VAE that compresses frames while preserving temporal cues.
Training departs from images along several axes, which we address in \S\ref{sec:method}:
(i) \emph{Compute and storage.} Longer sequences multiply the cost of sampling timesteps, noise draws, and frames, motivating fixed-timestep or small-subset estimators that reduce variance without prohibitive cost (\S\ref{sec:method:1}).
(ii) \emph{Variable horizon.} Clips vary in $\numFrames$ and frame rate (\S\ref{sec:method:video-bias}).
(iii) \emph{Time-specific failure modes.} Typical artifacts include inconsistent trajectories, temporal flicker, identity drift, and physically implausible dynamics despite sharp individual frames (\S\ref{sec:method:2}).

\textbf{Motion representations in videos.}
We denote our video as \(\videoclip=[\videoFrame_{\frameIndex}]_{\frameIndex=1}^{\numFrames}\) with \(\videoFrame_{\frameIndex}\in\mathbb{R}^{\videoHeight\times\videoWidth\times 3}\) being the 
$\frameIndex$-th frame.
We represent motion via optical flow between consecutive frames:
\(\motionFlow_{\frameIndex}:\{1,\ldots,\videoHeight\}\times\{1,\ldots,\videoWidth\}\to\mathbb{R}^2\), where each flow vector in $\mathbb{R}^2$ encodes the horizontal displacement $\mathrm{d}w$ and vertical displacement $\mathrm{d}h$ of a pixel. 
The motion magnitude is
\(\motionmagnitude(\videoHeightIndex,\videoWidthIndex)=\|\motionFlow_{\frameIndex}(\videoHeightIndex,\videoWidthIndex)\|_2\).
The \(\motionmagnitude\) over frames \(\frameIndex\) and pixels \(\videoHeightIndex,\videoWidthIndex\) summarizes the amount and spatial layout of motion in a clip, which we use to provide masks in our motion-weighted loss in \S\ref{sec:method}.

\providecommand{\annleftshift}{-3.0em}
\providecommand{\annrightshift}{ 3.0em}
\subsection{Data Attribution}
\label{subsec:attribution}

Data attribution measures how individual training samples affect a model's predictions~\citep{bae2024training}.
A classic approach to data attribution is to use influence functions~\citep{koh2017understanding}. Intuitively, the influence of a training sample measures: if we upweight this training example, how much would the model's prediction on a test datum change? 
Consider a loss function $\loss(\modelparams;\inputData)$ and a test sample $\inputData_{\textnormal{test}}$, the influence of a training point $\inputData_{\datasetIndex}$ can be quantified as:
\begin{align}
\influenceFunc(\inputData_{\datasetIndex},\inputData_{\textnormal{test}})
&~=~
- \nabla_{\modelparams}\loss(\modelparams;\inputData_{\textnormal{test}})^{\!\top}
\,\hessianmatrix^{-1}\,
\nabla_{\modelparams}\loss(\modelparams;\inputData_{\datasetIndex}), \quad
\hessianmatrix
=\tfrac{1}{\datasetSize}\sum\nolimits_{\datasetIndex=1}^{\datasetSize}
\nabla_{\modelparams}^{2}\loss(\modelparams;\inputData_{\datasetIndex}),
\end{align}
where the inverse Hessian captures the curvature of the loss landscape, yet computing or storing it is infeasible at modern model and dataset scales. Thus, practical methods (e.g., TracIn~\citep{pruthi2020estimating} and TRAK~\citep{park2023trak}) approximate influence via gradient inner products or gradient feature projections. 

\textbf{Attribution in diffusion models.}
Diffusion training aggregates gradients over timesteps $\timestep$ and noise draws $\noisevector$, where gradient norms vary systematically with $\timestep$, producing a timestep bias where examples aligned with large-norm timesteps appear spuriously influential. Diffusion-ReTrac~\citep{xie2024data} reduces this bias by normalizing gradients and sub-sampling $\timestep$ and $\noisevector$ for influence. Let $\lossdiff$ denote the diffusion loss, and with the sampled-timestep-and-noise set $\timestepset$, we compute a cosine-style score for normalized test and train gradients:
\begin{figure}[H]
\begin{equation}
\influenceFunc_{\textnormal{diff}}(\inputData_{\datasetIndex},\inputData_{\textnormal{test}})
\!=\!
\eqnmarkbox[NavyBlue]{testgrad}{
\frac{1}{|\timestepset_{\textnormal{test}}|}\!
\sum_{\timestep\!, \noisevector \in\timestepset_{\textnormal{test}}}
\frac{\nabla_{\modelparams}\lossdiff(\modelparams;\inputData_{\textnormal{test}},\timestep,\noisevector)}
{\big\|\nabla_{\modelparams}\lossdiff(\modelparams;\inputData_{\textnormal{test}},\timestep,\noisevector)\big\|}
^{\!\top}}
\,\, \,\, 
\eqnmarkbox[OliveGreen]{traingrad}{
\frac{1}{|\timestepset_{n}|}\!
\sum_{\timestep\!, \noisevector \in\timestepset_{n}}
\frac{\nabla_{\modelparams}\lossdiff(\modelparams;\inputData_{\datasetIndex},\timestep,\noisevector)}
{\big\|\nabla_{\modelparams}\lossdiff(\modelparams;\inputData_{\datasetIndex},\timestep,\noisevector)\big\|}}.
\label{eq:diffusion-influence}
\end{equation}

\annotate[xshift=\annleftshift ,yshift=-0.0em]{below,left }{testgrad}
{\scriptsize{normalized test gradients} \hspace{-0.055\textwidth}}
\annotate[xshift=\annrightshift,yshift=-0.1em]{below,left}{traingrad}
{\scriptsize{normalized training gradients \hphantom{aaaaaaaaa}}}
\end{figure}
\vspace{-0.025\textheight}
Averaging gradients over $(\timestep,\noisevector)$ stabilizes estimates, and normalization mitigates timestep-induced scale effects. Attribution quality is also sensitive to the measurement function used to score examples, such as denoising loss versus likelihood proxies~\citep{zheng2023intriguing}. 

\textbf{Why vanilla attribution is insufficient for videos.}
Na\"ively applying gradient-based attribution to video diffusion risks treating appearance and motion alike, overemphasizing appearance (objects, textures, backgrounds) while overlooking dynamics~\citep{park2025concept, tulyakov2018mocogan}. Its cost grows with clip length, sampled timesteps, noise draws, and gradient dimensionality, making na\"ive methods impractical at modern video scales. 
To improve motion, we need attribution that suppresses static appearance, emphasizes motion-specific signals, and remains efficient (\S\ref{sec:method}). Motion is distributed across frames and entangled with static cues, thus influence cannot be assigned frame-independently.

\vspace{-10pt}
\section{Method}
\label{sec:method}

We formalize the problem in \S\ref{sec:formulation} and develop a practical framework for motion attribution in video diffusion models with four components: scalable gradient computation (\S\ref{sec:method:1}), frame-length bias fix (\S\ref{sec:method:video-bias}), motion-aware weighting (\S\ref{sec:method:2}), and target data selection (\S\ref{sec:method:3}). We provide efficiency analysis (\S\ref{sec:method:4}) demonstrating scalability to billion-parameter models and large-scale video datasets.

\begin{figure*}[t]
\centering
\includegraphics[width=\textwidth]{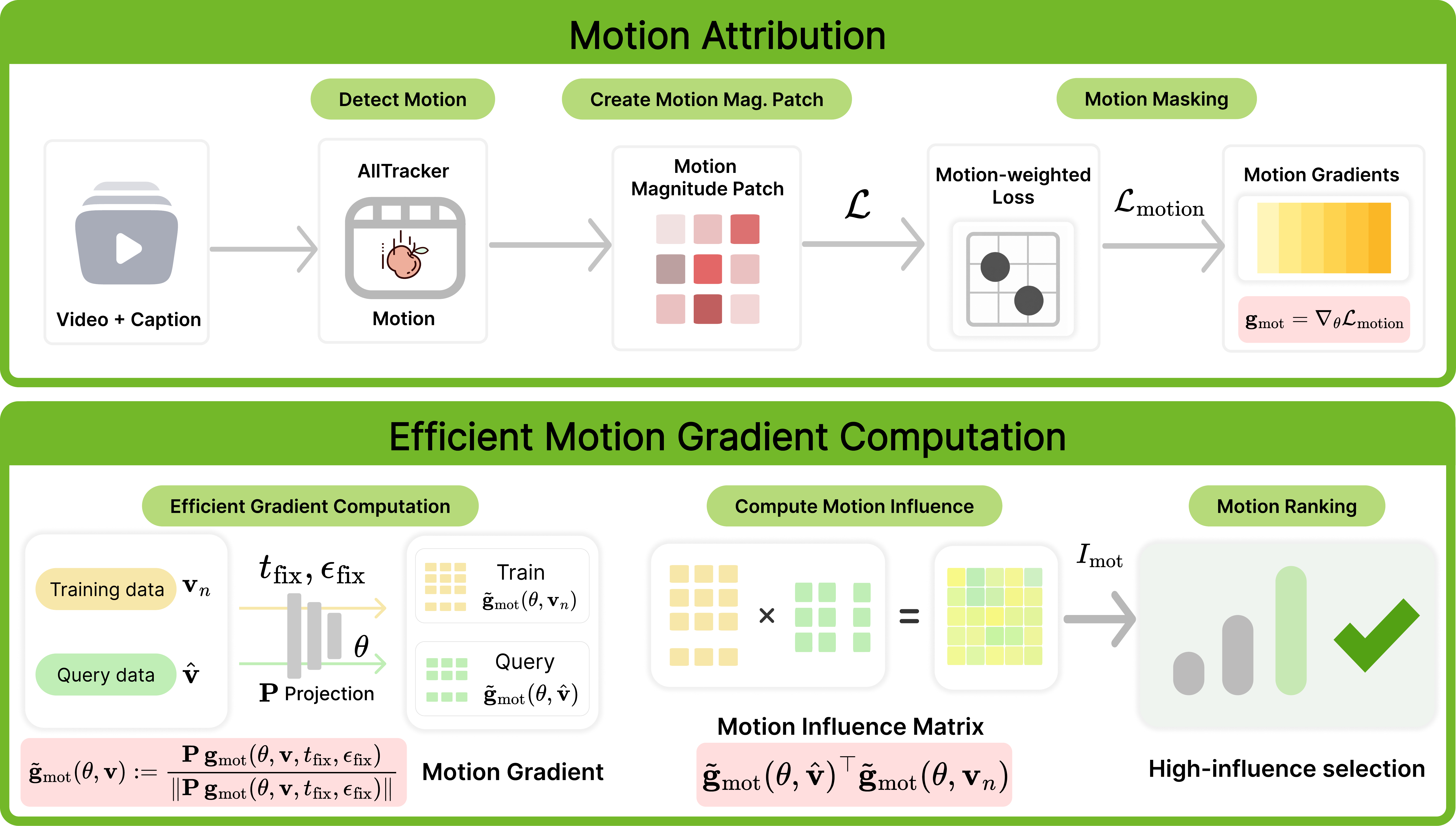}
\caption{\textbf{\methodname.}
\textit{Top.} Motion-gradient computation (\S\ref{sec:method:2}) has three steps: (1) detect motion with AllTracker, (2) compute motion-magnitude patches, (3) apply loss-space motion masks to focus gradients on dynamic regions. 
\textit{Bottom.} Our method (\S\ref{sec:method:1}) is made scalable via a single-sample variant with common randomness and a projection, computed for each pair of training and query data, aggregated (\S\ref{sec:method:3}) for a final ranking, and eventually used to select fine-tuning subsets.} 
\vspace{-10pt}
\label{fig:pipeline}
\end{figure*}

\subsection{Problem Formulation}
\label{sec:formulation}

We study data attribution for motion in the fine-tuning setting. Let
\(
\datasetft = \{(\videoclip_{\datasetIndex}, \conditioning_{\datasetIndex})\}_{\datasetIndex=1}^{\datasetSize}
\)
be the fine-tuning corpus. Given a query video \((\queryVideoclip,\queryConditioning)\), we assign to each training clip \((\videoclip_{\datasetIndex},\conditioning_{\datasetIndex})\) a motion-aware influence score
\(
\influenceFunc(\videoclip_{\datasetIndex}, \queryVideoclip; \modelparams)
\)
that explains how it contributes to the dynamics observed in \(\queryVideoclip\). The score should satisfy: 
(i) \emph{predictivity}, rankings correlate with observed changes from fine-tuning on the most influential subsets; 
(ii) \emph{efficiency}, scales to modern video generators, such as forgoing explicit Hessian inversion, expensive per-data integration, or prohibitive storage.
To do this, we augment the influence target defined in ~\eqref{eq:diffusion-influence} to be (a) lower variance for stable rankings with feasible levels of compute, (b) more scalable to store, and (c) motion-centric.

\textbf{Fine-tuning Subset Selection.}
For a budget \(\influenceSubsetSize\ll\datasetSize\), we get a motion-influential subset by ranking scores and taking the top-$\influenceSubsetSize$ examples. For multiple query motions, we combine selections as described in \S\ref{sec:method:3}. The resulting subsets serve as candidates for motion-centric fine-tuning.

\subsection{Scalable Gradient-based Attribution}
\label{sec:method:1}
We make attribution scalable for modern, large video datasets and models via inverse-Hessian approximations, lower-variance gradient-similarity estimators, low-cost single-sample estimators, and a Fastfood projection for tractable storage.

\textbf{Approximating the inverse-Hessian.}
Computing exact inverse-Hessian-vector products is infeasible for modern neural networks.
We estimate influence via gradient similarity, using an identity preconditioner for the inverse Hessian \citep{koh2017understanding,pruthi2020estimating,park2023trak}.

\textbf{Common randomness for stable rankings.}
To reduce variance without changing the target, we evaluate train and test gradients under the same $(\timestep,\noisevector)$ pairs and average over a small set $\timestepset$ \citep{xie2024data,lin2024diffusion}. This paired averaging stabilizes rankings compared to independent draws:
\vspace{-0.015\textheight}
\begin{figure}[H]
\begin{equation}
\label{eq:diffusion-influence-common}
\influenceFunc_{\textnormal{diff}}^{1}(\trainndata,\testdata)
=\frac{1}{|\timestepset|}
\sum_{\timestep,\noisevector\in\timestepset}
\eqnmarkbox[NavyBlue]{testgrad}{
\frac{\nabla_{\modelparams}\lossdiff(\modelparams;\,\testdata,\timestep,\noisevector)}
{\big\|\nabla_{\modelparams}\lossdiff(\modelparams;\,\testdata,\timestep,\noisevector)\big\|}
^{\!\top}}
\hspace{0.03\textwidth}
\eqnmarkbox[OliveGreen]{traingrad}{
\frac{\nabla_{\modelparams}\lossdiff(\modelparams;\,\trainndata,\timestep,\noisevector)}
{\big\|\nabla_{\modelparams}\lossdiff(\modelparams;\,\trainndata,\timestep,\noisevector)\big\|}}.
\end{equation}
\annotate[xshift=\annleftshift ,yshift=-0.0em]{below,left }{testgrad}{\scriptsize{normalized test gradients} \hspace{-0.055\textwidth}}
\annotate[xshift=\annrightshift,yshift=-0.1em]{below,left}{traingrad}{\scriptsize{normalized training gradients \hphantom{aaaaaaaaa}}}
\end{figure}
\vspace{-0.025\textheight}

\newpage
\textbf{Single-sample variant for reduced compute.}
We fix a single $\timestepfixed$ and a shared draw $\fixednoise\sim\mathcal{N}(0,\identity)$ for all train–test pairs. Sharing $(\timestepfixed,\fixednoise)$ allows low enough variance for the efficient single-sample estimator to maintain relative ordering~\citep{xie2024data,lin2024diffusion}. The estimator becomes:

\vspace{-0.015\textheight}
\begin{figure}[H]
\begin{equation}
\label{eq:diffusion-influence-single}
\influenceFunc_{\textnormal{diff}}^{2}(\trainndata,\testdata)
=
\eqnmarkbox[NavyBlue]{testgradS}{
\frac{\nabla_{\modelparams}\lossdiff(\modelparams;\,\testdata,\timestepfixed,\fixednoise)}
{\big\|\nabla_{\modelparams}\lossdiff(\modelparams;\,\testdata,\timestepfixed,\fixednoise)\big\|}
^{\!\top}}
\;\;
\eqnmarkbox[OliveGreen]{traingradS}{
\frac{\nabla_{\modelparams}\lossdiff(\modelparams;\,\trainndata,\timestepfixed,\fixednoise)}
{\big\|\nabla_{\modelparams}\lossdiff(\modelparams;\,\trainndata,\timestepfixed,\fixednoise)\big\|}}.
\end{equation}
\annotate[xshift=\annleftshift ,yshift=-0.0em]{below,left }{testgradS}{\scriptsize{normalized test gradient} \hspace{-0.055\textwidth}}
\annotate[xshift=\annrightshift,yshift=-0.1em]{below,left}{traingradS}{\scriptsize{normalized training gradient} \hphantom{aaaaaaaaa}}
\end{figure}
\vspace{-0.025\textheight}

\textbf{Structured projection for reduced storage.}
To operate at model scale, we apply a Johnson–Lindenstrauss projection via Fastfood \citep{le2014fastfood} and then normalize. 
Let
\vspace{-5pt}
\begin{equation}
\projectionMatrix \in \mathbb{R}^{\projecteddim \times \gradientdim}
\;\;\text{implemented as}\;\;
\projectionMatrix :=\tfrac{1}{\fastfoodNorm\sqrt{\projecteddim}}\, \mathbf{S} \mathbf{Q} \mathbf{G} \boldsymbol{\Pi} \mathbf{Q} \mathbf{B},
\vspace{-5pt}
\end{equation}
where $\mathbf{Q}$ is the Walsh–Hadamard matrix, $\mathbf{B}$ is a diagonal Rademacher matrix, $\boldsymbol{\Pi}$ is a random permutation, $\mathbf{G}$ is a diagonal Gaussian scaling, and $\mathbf{S}$ is a diagonal rescaling, and $\fastfoodNorm$ normalizes the variance. The projected, normalized gradient is:
\vspace{-5pt}
\begin{equation}
\gradientProjected\big(\modelparams, \inputData\big)
~:=~
\frac{
\projectionMatrix\,\nabla_{\modelparams}\lossdiff(\modelparams, \inputData,\timestepfixed,\fixednoise)
}{
\|\projectionMatrix\,\nabla_{\modelparams}\lossdiff(\modelparams, \inputData,\timestepfixed,\fixednoise)\|
}.
\end{equation}
Then the influence score is the dot product of normalized projected gradients (i.e., cosine similarity) in $\mathbb{R}^{\projecteddim}$:
\vspace{-0.015\textheight}
\begin{equation}
\label{eq:diffusion-influence-tilde}
\influenceFunc_{\textnormal{diff}}^{3}(\trainndata,\testdata)
~=~
\eqnmarkbox[NavyBlue]{tildegtest}{
\gradientProjected\!\big(\modelparams;\,\testdata\big)^{\!\top}}
\;\;
\eqnmarkbox[OliveGreen]{tildegtrain}{
\gradientProjected\!\big(\modelparams;\,\trainndata\big)}.
\end{equation}
\annotate[xshift=\annleftshift ,yshift=-0.0em]{below,left }{tildegtest}{\scriptsize{projected, normalized test gradient} \hspace{-0.055\textwidth}}
\annotate[xshift=\annrightshift,yshift=-0.1em]{below,right}{tildegtrain}{\hspace{-0.06\textwidth}\scriptsize{projected, normalized training gradient}}
\vspace{-0.02\textheight}

This keeps compute $\mathcal{O}(\projecteddim\log\projecteddim)$ for projection and $\mathcal{O}(\projecteddim)$ per dot product, with storage $\mathcal{O}(|\dataset|\,\projecteddim)$, while staying close to the ranking behavior of full-gradient cosine similarity \citep{park2023trak}.

\subsection{Video-specific Frame-length Bias Fix}
\label{sec:method:video-bias}
Raw gradient magnitudes depend on the number of frames $\numFrames$ in the video $\videoclip$, thereby biasing scores toward longer videos. 
We correct this by normalizing for frame count before the projection–normalization step:
\begin{equation}
\smash{
\nabla_{\modelparams}\lossdiff(\modelparams;\,\videoclip,\timestepfixed,\fixednoise)
\;\leftarrow\;
\frac{1}{\numFrames}\,
\nabla_{\modelparams}\lossdiff(\modelparams;\,\videoclip,\timestepfixed,\fixednoise).
}
\end{equation}
We still apply $\ell_2$ normalization in \eqref{eq:diffusion-influence-tilde}, further stabilizing scales across examples. Together, single-timestep, common randomness, projection, and frame-length correction form a compact, scalable estimator that we use throughout. 
However, na\"ive video-level attribution conflates appearance with motion, often ranking clips high due to shared backgrounds or objects, while offering little insight into dynamics.

\vspace{-10pt}
\subsection{Motion Attribution}
\label{sec:method:2}
To move beyond whole-video influence, we introduce motion attribution, which isolates the contribution of training data to temporal dynamics. Unlike video-level attribution, which treats each clip as a single unit and conflates appearance with motion, motion attribution reweights per-location gradients using motion masks, assigning influence via dynamic behavior rather than static content.

\textbf{Motion Masking Attribution.}
Motion is what distinguishes video diffusion from image diffusion. 
Our goal is to understand how training data shapes motion in video diffusion models.
Prior work has emphasized architectural or algorithmic changes for motion modeling~\citep{peebles2023scalable, blattmann2023stable, guo2023animatediff}, many of the largest generative gains have instead come from scaling and curating massive video corpora, which in turn enable impressive motion synthesis results in video diffusion models~\citep{ho2022video, wan2025, tan2024vidgen, yang2024cogvideox}. Yet we lack tools that quantify how specific training clips shape particular motion patterns. We address this by attributing motion back to data via motion-weighted gradients, which yields actionable signals for targeted data selection, artifact diagnosis, and selective fine-tuning.

\textbf{Motion Detection and Latent Space Mapping.} 
Given a video $\videoclip \in \mathbb{R}^{\numFrames \times \videoHeight \times \videoWidth \times 3}$ with $\numFrames$ frames of resolution $\videoHeight \times \videoWidth$, we first encode it into the VAE latent space as $\latents = \vaeencoder(\videoclip) \in \mathbb{R}^{\numFrames \times \nicefrac{\videoHeight}{\latentDownsampleFactor} \times \nicefrac{\videoWidth}{\latentDownsampleFactor} \times \latentChannelSize}$, with downsampling factor $\latentDownsampleFactor = 8$ and $\latentChannelSize = 16$ following the \sbt{Wan2.1} backbone used in our experiments.

For motion computation, we use AllTracker~\citep{harley2025alltracker} to extract motion information in pixel space: $\motioninfo = \alltracker(\videoclip) \in \motionspace$, where the first two channels contain optical flow maps $\motioninfo_{:,:,:,0:2}$ indicating pixel displacement between frames, and the remaining channels $\motioninfo_{:,:,:,2:4}$ encode visibility and confidence scores. We extract displacement vectors at each pixel location as:\vspace{-0.01\textheight}
\begin{equation}
\label{eq:displacement-vector}
\smash{
\displacementvec(\videoHeightIndex,\videoWidthIndex) = (\motioninfo_{\frameIndex,\videoHeightIndex,\videoWidthIndex,0}, \motioninfo_{\frameIndex,\videoHeightIndex,\videoWidthIndex,1}) = (\mathrm{d}w, \mathrm{d}h).
}\vspace{-0.01\textheight}
\end{equation}
We then bilinearly downsample motion quantities from $(\videoHeight,\videoWidth)$ to the latent grid $\big(\tfrac{\videoHeight}{\latentDownsampleFactor},\tfrac{\videoWidth}{\latentDownsampleFactor}\big)$ so that our masking lives where gradients are computed.

\textbf{Motion-Weighted Gradient Computation.}
We define the motion magnitude at each location as: $\motionmagnitude(\videoHeightIndex,\videoWidthIndex)
\!=\!\|\displacementvec(\videoHeightIndex,\videoWidthIndex)\|_2$.
To obtain comparable motion weights across frames and pixels, we min–max normalize over all frames and pixels, ensuring values lie in $[0,1]$:\vspace{-0.005\textheight}
\begin{equation}
\label{eq:motion-weight}
\motionweights(\frameIndex,\videoHeightIndex,\videoWidthIndex)
=
\frac{\motionmagnitude(\videoHeightIndex,\videoWidthIndex)-m_{\min}}
{m_{\max}-m_{\min}+\numericalBias},
\end{equation}
where $m_{\min}=\min_{\frameIndex',\videoHeightIndex',\videoWidthIndex'}\motionmagnitude(\videoHeightIndex',\videoWidthIndex')$, $m_{\max}=\max_{\frameIndex',\videoHeightIndex',\videoWidthIndex'}\motionmagnitude(\videoHeightIndex',\videoWidthIndex')$, and $\numericalBias=10^{-6}$ ensures a positive denominator.  
This normalization mitigates bias from absolute motion scale, yielding weights that emphasize relative motion saliency rather than raw magnitude, following prior practice in video saliency detection~\citep{fang2013video}.
Let $(\latentHeightIndex,\latentWidthIndex)$ index the latent grid. We obtain latent-aligned weights by bilinear downsampling:
\begin{equation}
\label{eq:motion-downsample}
\smash{
\tilde{\motionweights}(\frameIndex,\latentHeightIndex,\latentWidthIndex)
~=~\operatorname{Bilinear}\!\left(\motionweights(\cdot, \cdot,\cdot),\numFrames,~\tfrac{\videoHeight}{\latentDownsampleFactor},~\tfrac{\videoWidth}{\latentDownsampleFactor}\right).
}
\end{equation}
We compute per-location squared error at fixed $(\timestepfixed,\fixednoise)$ at each frame $\frameIndex$ and ``latent pixel'' $(\latentHeightIndex,\latentWidthIndex)$:
\vspace{-5pt}
\begin{equation}
\label{eq:per-location-error}
\begin{aligned}
\perLocError_{\modelparams,\videoclip,\conditioning}(\frameIndex, \latentHeightIndex,\latentWidthIndex)
~=~\Big(\,&[\noisepredict(\noisyLatents(\videoclip, \timestepfixed, \fixednoise),\timestepfixed,\conditioning)]_{\frameIndex,\latentHeightIndex,\latentWidthIndex} 
- [\trainingtarget(\timestepfixed,\fixednoise)]_{\frameIndex,\latentHeightIndex,\latentWidthIndex}\,\Big)^2,
\end{aligned}
\end{equation}
and define the motion-weighted loss by averaging over frames and latent spatial locations:
\begin{equation}
\label{eq:motion-loss}
\lossmotionweighted(\modelparams;\videoclip,\conditioning)
~=~\frac{1}{\numFrames_{\videoclip}} \operatorname*{mean}_{\frameIndex,\latentHeightIndex,\latentWidthIndex} 
\Big[\tilde{\motionweights}_{\videoclip,\conditioning}(\frameIndex,\latentHeightIndex,\latentWidthIndex)\cdot \perLocError_{\modelparams,\videoclip,\conditioning}(\frameIndex, \latentHeightIndex,\latentWidthIndex)\Big].
\end{equation}
Notably, when $\tilde{\motionweights}$ is all ones, this recovers the standard objective with no motion emphasis. The $\nicefrac{1}{\numFrames_{\videoclip}}$ factor corrects for frame-length bias and $\numFrames_{\videoclip}$ signifies how the number of frames may be video-dependent. The corresponding motion-weighted gradient for attribution is:
\begin{align}
\label{eq:diffusion-influence-motion}
\InfluenceMot\!(\trainVideoData,\!\queryVideoclip)
&\!=\!
\gradientMotionProjected\!(\modelparams\!,\!\queryVideoclip)^{\!\top}\!
\gradientMotionProjected\!(\modelparams\!,\!\trainVideoData), \quad
\textnormal{where } \gradientMotion \!:=\! \nabla_{\!\modelparams}\lossmotionweighted 
\textnormal{ and } \gradientMotionProjected\!(\modelparams\!,\!\videoclip) \!:=\!
\frac{
\projectionMatrix\gradientMotion\!(\modelparams\!,\!\videoclip\!,\!\timestepfixed,\!\fixednoise)
}{
\|\projectionMatrix\gradientMotion\!(\modelparams\!,\!\videoclip\!,\!\timestepfixed,\!\fixednoise)\|
}.
\end{align}

Loss-space masking leaves forward noising and generation unchanged and reweights only attribution, avoiding interactions between motion weighting and noise injection. In contrast, our motion-aware attribution emphasizes dynamic regions and de-emphasizes static backgrounds, so rankings identify training clips that most strongly shape the model's motion rather than appearance.

\subsection{Most Influential Fine-tuning Subset Selection}
\label{sec:method:3}

\textbf{Goal.} Given a query clip \((\queryVideoclip,\queryConditioning)\), we compute a motion-aware attribution
value for each fine-tuning sample \((\trainVideoData,\conditioning_{\datasetIndex}) \in \datasetft\) using $\InfluenceMot\!\left(\trainVideoData,\,\queryVideoclip\right)$ (\eqref{eq:diffusion-influence-motion}). We construct a fine-tuning set $\dataSubset$ for one or many query videos $\queryVideoclip$.

\textbf{Single-query-point fine-tuning selection.}
For a budget of $\influenceSubsetSize$ data points, we select the \(\influenceSubsetSize\) highest-scoring examples.
In practice, $\influenceSubsetSize$ is chosen as a percentile of the dataset size (e.g., top 1–10\%), ensuring the subset scales consistently across datasets.

\newpage
\textbf{Multi-query-point fine-tuning selection: aggregating attribution scores.}
For $\numMotions$ queries, we adopt the majority voting approach from ICONS~\citep{wu2024icons} and aggregate motion-aware influence scores across queries by percentile thresholding and voting. A sample receives a vote if the score is above the percentile cutoff $\influenceThreshold$ for that query. 
The consensus score of a candidate $\trainVideoData$ is the total number of queries that vote for it. We then rank all training samples by $\MajVote(\trainVideoData)$ and select the top-$\influenceSubsetSize$ to form the fine-tuning subset. This formulation emphasizes samples that are consistently influential across multiple queries, without requiring cross-query calibration of raw scores:
\begin{equation}
\label{eq:majority-vote}
\begin{aligned}
\MajVote_{\datasetIndex}
&\!=\!
\sum\nolimits_{\motionIndex=1}^{\numMotions}\!\mathbb{I}\!\big[\,
\InfluenceMot(\trainVideoData,\queryVideoclip_\motionIndex) \!>\! \influenceThreshold
\,\big], \quad
\dataSubset_{\textnormal{vote}}(\influenceSubsetSize)
\!=\!
\bigl\{\trainVideoData | \trainVideoData
\text{ in top-}\influenceSubsetSize \text{ by } \MajVote \bigr\}.
\end{aligned}
\end{equation}

\vspace{-10pt}
\subsection{Computational Efficiency Analysis}
\label{sec:method:4}

\textbf{Gradient Compute.} Na\"ively averaging over timesteps and noise for every example costs $\mathcal{O}(|\dataset|\,|\timestepset|\,\cost)$, where $\cost$ is a single forward+backward cost and $|\timestepset|$ is the number of sampled $\timestep, \noisevector$ per data. 
Using a single sample reduces this to $\mathcal{O}(|\dataset|\,\cost)$, which is key to keeping the cost reasonable for modern video datasets and models, while reusing a sample across data allows low enough variance for stable rankings. Projection adds $\mathcal{O}(\projecteddim \log \projecteddim)$ per example using Fastfood \citep{le2014fastfood}, negligible relative to a backward pass.

\textbf{Gradient Storage.} Storing full gradients is $\mathcal{O}(|\dataset|\,\gradientdim)$. We instead store only projected vectors, $\mathcal{O}(|\dataset|\,\projecteddim)$, plus the structured Fastfood state, $\mathcal{O}(\gradientdim)$. Since $\projecteddim$ is typically orders of magnitude smaller than $\gradientdim$, this transformation makes storage tractable for billion-parameter models.

\textbf{Data Ranking Compute.} Influence computation in ~\eqref{eq:diffusion-influence-tilde} is an inner product in $\mathbb{R}^{\projecteddim}$, so evaluating all train examples against a query is $\mathcal{O}(|\dataset|\,\projecteddim)$, and sorting is $\mathcal{O}(|\dataset|\log|\dataset|)$.

\begin{figure*}[t]
   \centering
   \includegraphics[width=\textwidth]{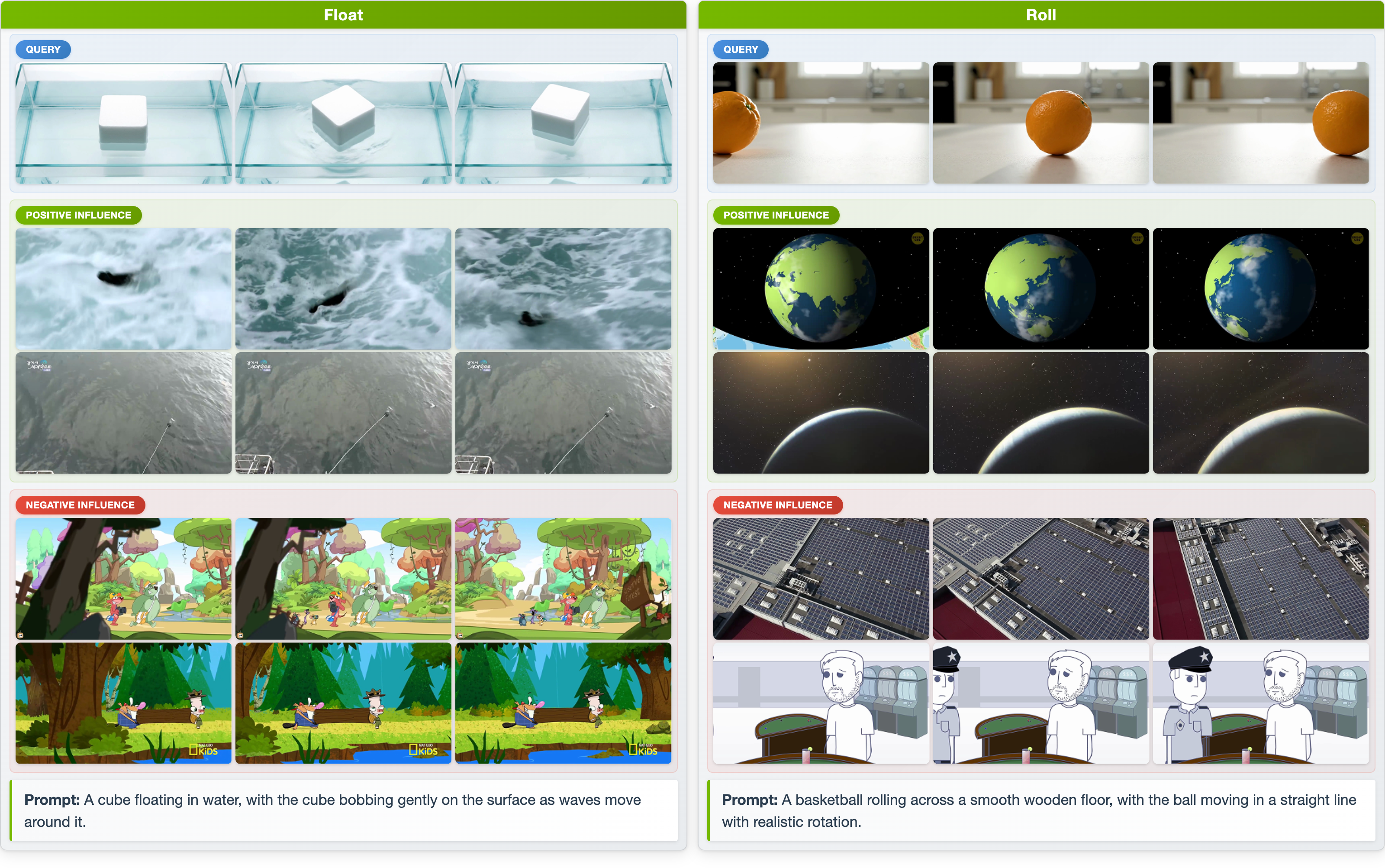}
   \vspace{-15pt}
   \caption{\textbf{Motion Attribution Samples} with Wan2.1-T2V-1.3B. 
   \textit{Top}: Query clips showing float (\textit{left}) and roll (\textit{right}) motions. 
   \textit{Middle}: 
   Top-ranked positive training samples identified by \methodname~with high influence scores.
   \textit{Bottom}: Negative influence samples with minimal, camera-only motion, or cartoon-style content that conflict with target motions. Videos are included in the supplementary material.}
   \vspace{-15pt}
   \label{fig:influence_examples}
\end{figure*}

\newpage
\textbf{Additional Motion-Emphasis Compute.} 
Motion-specific overhead primarily stems from AllTracker mask extraction with complexity $\mathcal{O}(|\dataset| \cdot \videoHeight \cdot \videoWidth \cdot \numFrames)$ for clip length $\numFrames$ and frame resolution $\videoHeight \times \videoWidth$. Masks are extracted once, cached, and negligible relative to gradient cost. We provide a detailed runtime breakdown in App.~\ref{app:runtime}.

\section{Experiment}
\label{sec:experiment}

\subsection{Setup}
\label{sec:experiment:setup}

\textbf{Fine-tuning Datasets.}
We evaluate our motion attribution framework on two large-scale video datasets: \sbt{VIDGEN-1M}~\citep{tan2024vidgen} and \sbt{4DNeX-10M}~\citep{chen20254dnex}, both of which offer diverse motion patterns, rich temporal dynamics, and complex scenes. For our experiments, we use 10k videos from both datasets, which provide sufficient scale and diversity to thoroughly evaluate motion attribution methods across different temporal patterns and video generation scenarios.

\textbf{Motion Query Data.}
To evaluate our motion attribution, we curate a set of query videos representing distinct motion patterns and scenarios. Our query dataset consists of videos spanning $10$ motion categories, with a focus on object dynamics: \emph{compress, bounce, roll, explode, float, free fall, slide, spin, stretch, swing}.
Five videos, totaling $50$ queries, represent each motion type. These videos are chosen for their clear, isolated motions, serving as a basis for evaluating attribution quality and downstream motion generation. Details are in App.~\ref{app:motion_query_data}.

\textbf{Model \& Baselines.}
We use pretrained models \sbt{Wan2.1-T2V-1.3B} and \sbt{Wan2.2-TI2V-5B}, with additional results on \sbt{LTX-2B} in App.~\ref{sec:add_experiments}. Our baselines: 
Base model: pretrained, no fine-tuning; 
Full fine-tuning: approximate upper bound using the complete dataset; 
Random selection: uniform sampling; 
Motion magnitude: selects videos with the highest average motion magnitude; 
V-JEPA embeddings: selects most representative videos of motion patterns using V-JEPA~\citep{assran2025v} features, capturing high-level motion semantics;
and Ours w/o motion masking: influence of the entire video level without motion-specific masking.

\begin{figure*}[!t]
\centering
\vspace{-14pt}
\includegraphics[width=\textwidth]{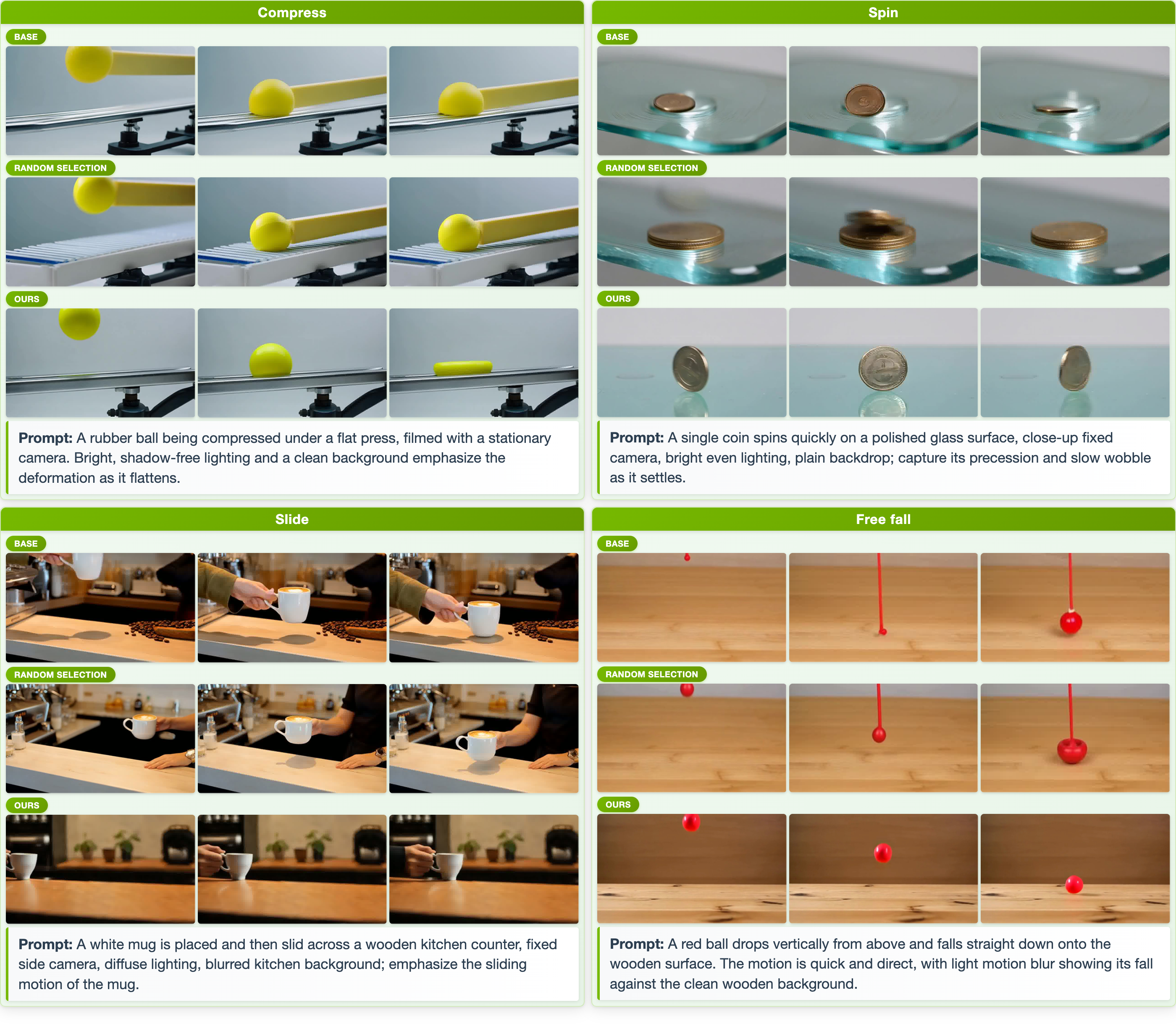}
\vspace{-20pt}
\caption{\textbf{Qualitative Comparisons.}
We compare four motion scenarios across the base model (Wan2.1-T2V-1.3B), or fine-tuned with random selection and \methodname. Our approach produces more realistic motion. Videos are included in the supplementary material.
}
\label{fig:qualitative_results}
\vspace{-5pt}
\end{figure*}

\begin{table*}[t!]
    \centering
    \caption{\textbf{VBench Evaluation.} Performance comparison on VBench~\citep{huang2024vbench} across different baselines (all values in $\%$, higher is better). All selection methods use $10\%$ of training data; our method uses majority vote aggregation (\S\ref{sec:method:3}) across motion queries. MM: motion masking. Subj.: Subject Consistency, Bg.: Background Consistency, Mot.: Motion Smoothness, Dyn.: Dynamic Degree, Aesth.: Aesthetic Quality, Img.: Imaging Quality.
    }
    \vspace{-5pt}
    \label{tab:vbench-wan}
    \small
    \renewcommand{\arraystretch}{1.12}
    \setlength{\tabcolsep}{4pt}
    \resizebox{0.95\textwidth}{!}{%
    \begin{tabular}{@{}l*{6}{r}@{\hspace{0.6em}\vrule\hspace{0.6em}}*{6}{r}@{}}
    \toprule
    & \multicolumn{6}{c}{Wan2.1-T2V-1.3B} & \multicolumn{6}{c}{Wan2.2-TI2V-5B} \\
    \cmidrule(lr){2-7} \cmidrule(lr){8-13}
    Method & Subj. & Bg. & Mot. & Dyn. & Aesth. & Img. & Subj. & Bg. & Mot. & Dyn. & Aesth. & Img. \\
    \midrule
    Base        & 95.3 & 96.4 & 96.3 & 39.6 & 45.3 & \textbf{65.7} & 94.9 & 96.4 & 97.5 & 42.0 & 44.4 & 65.5 \\
    Full FT     & 95.9 & \textbf{96.6} & 96.3 & 42.0 & 45.0 & 63.9           & \textbf{95.3} & 96.5 & 97.5 & 45.3 & 44.8 & \textbf{66.2} \\
    Random      & 95.3 & 96.6 & 96.3 & 41.3 & 45.7 & 65.1           & 94.7 & 96.2 & 97.3 & 41.6 & 44.6 & 65.2 \\
    Motion mag. & 95.6 & 96.2 & 95.7 & 40.1 & 45.1 & 63.2           & 95.0 & 96.3 & 97.4 & 44.9 & 45.0 & 65.1 \\
    V-JEPA      & 95.7 & 96.0 & 95.6 & 41.6 & 44.9 & 62.7           & 95.2 & 96.4 & 97.3 & 45.6 & 44.9 & 64.8 \\\midrule
    Ours w/o MM & 95.4 & 96.1 & 96.3 & 43.8 & 45.7 & 63.2           & 94.9 & 96.5 & 97.4 & 43.8 & 45.2 & 64.8 \\
    \rowcolor{ggreen!10}
    \textbf{Ours (\methodname)} & \textbf{96.3} & 96.1 & 96.3 & \textbf{47.6} & \textbf{46.0} & 64.6 & 95.1 & \textbf{96.6} & \textbf{97.6} & \textbf{48.3} & \textbf{45.6} & 65.5 \\
    \bottomrule
    \end{tabular}
    }
    \vspace{-0pt}
\end{table*}

\textbf{Benchmark.}
We evaluate our attribution using VBench~\citep{huang2024vbench} metrics across six dimensions: subject consistency, background consistency, motion smoothness, dynamic degree, aesthetic quality, and imaging quality. Motion smoothness and dynamic degree are our primary targets for temporal dynamics, while other metrics help maintain visual quality. We use custom evaluation prompts, following VBench's descriptive style, for 10 motion types, with $5$ prompts each, to assess our framework's effectiveness on target motions.

\newpage
\textbf{Implementation Details.}
We finetune base models with our \methodname-selected high-quality video data following the official and DiffSynth-Studio implementation. 
During fine-tuning, we update only the DiT backbone while freezing the T5 text encoder and VAE. All models are trained at a resolution of $480\times832$ with a learning rate of $1 \times 10^{-5}$. 
Specialist models are trained on single motion category selected data, while generalist models use aggregated selections (both with top $10\%$ selection from \sbt{VIDGEN-1M}~\citep{tan2024vidgen} or \sbt{4DNeX}~\citep{chen20254dnex}). 
Compute and runtime details are provided in App.~\ref{app:runtime}.

\subsection{Main Results}
\label{sec:main_results}

\textbf{High-influence selection and negative filtering.}
Fig.~\ref{fig:influence_examples} shows our motion-aware attribution ranks clips with clear, physically grounded dynamics and downranks those with little transferable motion. For rolling and floating, positives show continuous trajectories and smooth temporal evolution (turbulent water, planetary rotation). Negatives are mostly static footage, camera-only motion, or cartoons whose simplified kinematics do not transfer. This pattern holds across motion categories and aligns with the quantitative gains below.

\textbf{Qualitative improvements across motion types.}
Fig.~\ref{fig:qualitative_results} compares the base \sbt{Wan2.1-T2V-1.3B} model, finetuned with random selection or \methodname with equal data budgets, across $4$ scenarios. 
Our method yields higher motion fidelity and temporal consistency than baselines, especially for complex deformation and physics-driven motion.

\textbf{Quantitative Results.} 
We evaluate our approach using VBench~\citep{huang2024vbench}, demonstrating consistent improvements in motion when fine-tuning with attribution-selected data.
As shown in Tab.~\ref{tab:vbench-wan}, \methodname~consistently achieves the highest dynamic degree scores across both models: $47.6\%$ on \sbt{Wan2.1-T2V-1.3B} and $48.3\%$ on \sbt{Wan2.2-TI2V-5B}, significantly outperforming random selection ($41.3\%$ and $41.6\%$) and whole video attribution ($43.8\%$ for both models). Our method also excels in aesthetic quality ($46.0\%$ and $45.6\%$), while maintaining competitive motion smoothness ($96.3\%$ and $97.6\%$). Notably, using only $10\%$ of training data, our approach surpasses the full fine-tuning on dynamic degree ($42.0\%$ and $45.3\%$) on both models, demonstrating superior performance of motion-specific attribution for targeted fine-tuning. Additional results on \sbt{LTX-2B} are in App.~\ref{sec:add_experiments}; motion magnitude distribution analysis is in App.~\ref{sec:analysis}.

\begin{figure*}[!t]
\centering
\includegraphics[width=\textwidth]{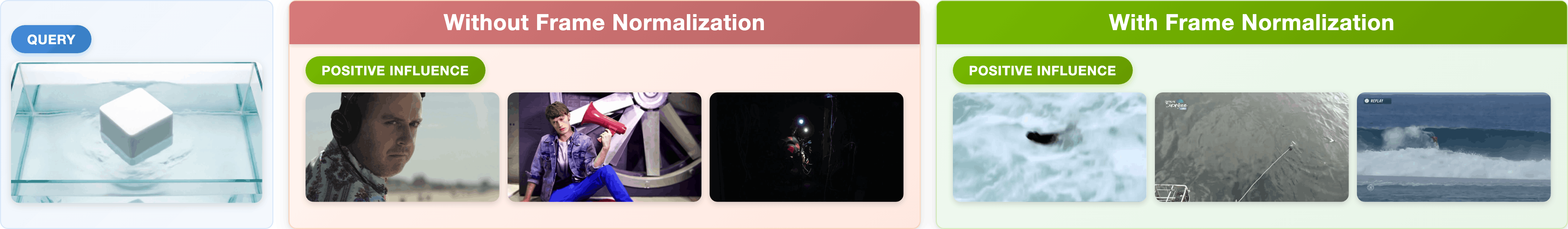}
\caption{\textbf{Impact of Frame-Length Normalization on Motion Attribution.} Comparison of top-ranked samples for floating motion query. \textbf{Left}: With proper frame-length normalization, top samples consistently exhibit floating motion (waves, floating objects, surfing). \textbf{Right}: Without normalization, rankings are biased by video length, resulting in no coherent patterns among top samples. 
}
\label{fig:frame_norm}
\vspace{-15pt}
\end{figure*}

\subsection{Human Evaluation}

\begin{wraptable}[10]{r}{0.45\textwidth}
    \centering
    \vspace{-0.02\textheight}
    \small
    \begin{tabular}{@{}lccc@{}}
    \toprule
    \textbf{Method} & \textbf{Win ($\%$)} & \textbf{Tie ($\%$)} & \textbf{Loss ($\%$)} \\
    \midrule
    Ours vs. Base & $74.1$ & $12.3$ & $13.6$ \\  %
    Ours vs. Random & $58.9$ & $12.1$ & $29.0$ \\  %
    Ours vs. Full FT & $53.1$ & $14.8$ & $32.1$ \\  %
    Ours vs. w/o MM & $46.9$ & $20.0$ & $33.1$ \\  %
    \bottomrule
    \end{tabular}
    \vspace{-3pt}
    \caption{\textbf{Human evaluation.} Pairwise comparisons across $50$ videos with $17$ participants ($850$ total). Win, tie, and loss rates show where our method is preferred, rated equal, or outperformed.}
    \label{tab:human_eval_extended}
\end{wraptable}

Automated scores can miss perceptual motion quality, so we run a human evaluation pairwise comparison protocol: participants view two generated videos and choose which shows better motion. We recruit $17$ annotators and evaluate $10$ motion categories. For each category, we prepare $5$ test cases and compare our method to baselines across three pairings, yielding a balanced set of judgments. Presentation order is randomized, and ties are allowed. We report win rate (fraction our method is preferred), tie rate, and overall preference. As shown in Tab.~\ref{tab:human_eval_extended}, annotators favor our attribution-guided selection: $74.1\%$ win rate vs. the base model and $53.1\%$ vs. the full fine-tuned model, showing perceptually meaningful motion improvements.

\subsection{Ablations}
\label{subsec:ablations}

\begin{wrapfigure}{r}{0.45\textwidth}
\centering
\vspace{-0.025\textheight}
\begin{tikzpicture}
    \centering
    \node (img11) {\includegraphics[width=.9\linewidth]{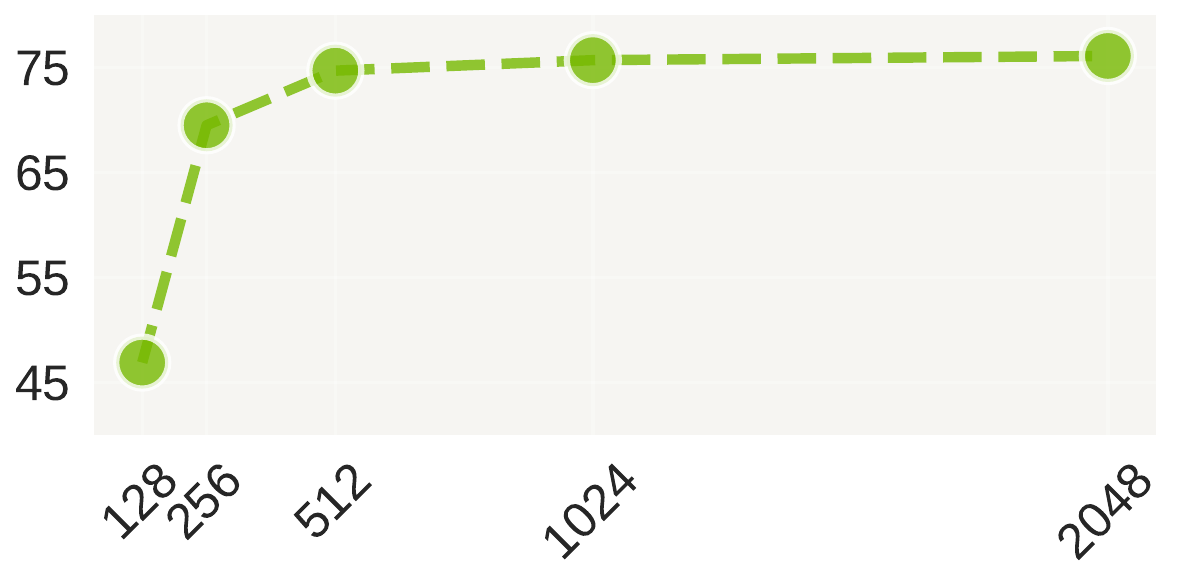}};

    \node[left=of img11, rotate=90, anchor=center, xshift=0.2cm, yshift=-1.0cm] {\footnotesize{Spearman Correlation $\spearmancorr$}};
    \node[below=of img11, node distance=0cm, xshift=0.1cm, yshift=1.3cm, font=\color{black}] {\footnotesize{Projection Dimension}};
\end{tikzpicture}
\caption{\textbf{Projection dimension analysis.} Spearman correlation between projected and full gradients shows rapid improvement with projection dimension, with $512$ providing a strong trade-off between accuracy and efficiency.}
\label{fig:projection_correlation}
\vspace{-15pt}
\end{wrapfigure}

\textbf{Projected Gradients Preserve Influence Rankings.} Comparing full gradients for attribution is infeasible at a billion-parameter scale. We reduce dimensionality with structured random projections that preserve influence geometry, ablating $\projecteddim \!\in\! \{128,\dots,2048\}$ against the full-gradient baseline. We assess ranking preservation via Spearman correlation with unprojected scores (Fig.~\ref{fig:projection_correlation}). Small projections preserve rankings poorly: $\projecteddim\!=\!128$ yields $\spearmancorr\!=\!46.9\%$. Preservation improves with size: $\projecteddim\!=\!512$ reaches $\spearmancorr\!=\!74.7\%$. Beyond that, gains are marginal while cost rises: $\projecteddim\!=\!1024$ ($\spearmancorr\!=\!75.7\%$) and $\projecteddim\!=\!2048$ ($\spearmancorr\!=\!76.1\%$). Thus, $\projecteddim\!=\!512$ offers the best trade-off, scaling to large models while maintaining quality.

\newpage
\textbf{Single-timestep attribution.}
Using a single timestep avoids the cost of averaging across timesteps while closely matching the multi-timestep baseline. With fixed $\timestep\!=\!751$ (midpoint of the $1000$-step trajectory), we obtain $\spearmancorr\!=\!66\%$ agreement with ground truth computed using $10$ evenly-spaced timesteps based on flow matching schedule.
High timesteps (early denoising) heavily obscure motion with noise; low timesteps (late denoising) operate on nearly formed videos, where gradients reflect fine details. $\timestep\!=\!751$ (mid-denoising)  balances influence ranking correlation and compute efficiency. Averaging multiple timesteps yields minimal gains, a single fixed timestep is therefore sufficient for variance-reduced, scalable attribution.

\textbf{Frame-Length Normalization.}
As in the Wan training protocol, we standardize all videos to 81 frames at 16 fps (satisfying the $4n{+}1$ constraint) for fair attribution across clips of different raw lengths. Without standardization, gradient-based scores correlate strongly with video length rather than motion quality ($\spearmancorr\!=\!78.0\%$), leading to longer clips ranking higher regardless of dynamics. Standardizing frames reduces spurious length correlations by $54.0\%$ while preserving motion-based correlation, so rankings reflect motion rather than duration.
As in Fig.~\ref{fig:frame_norm}, normalization clarifies motion-specific patterns. For floating queries with frame-length normalization (left), top-ranked samples consistently show wave dynamics, floating objects, and surfing, all matching the target motion. Without normalization (right), top samples lack coherent similarity because rankings are driven by clip length, harming motion-relevant training example identification.

\section{Conclusion}

We address a central and underexplored question in video diffusion: which training clips influence the motion in generated videos? We introduced \methodname, a motion-aware data attribution framework for video diffusion that isolates temporal dynamics from static appearance. By tracing generated motion back to influential training clips, our method enables targeted data curation that improves motion quality with a fraction of the data. As video models scale, such data-level understanding will be key to diagnosing failure modes and building more controllable generative systems.

\section*{Acknowledgements}
We thank the following people (listed alphabetically by last name) for their helpful discussions, feedback, or participation in human studies: Allison Chen, Sanghyuk Chun, Zhiwei Deng, Amaya Dharmasiri, Xingyu Fu, Will Hwang, Yifeng Jiang, Amogh Joshi, Pang Wei Koh, Chen-Hsuan Lin, Huan Ling, Tiffany Ling, Shaowei Liu, Zhengyi Luo, Rafid Mahmood, Kaleb S. Newman, Julian Ost, Zeeshan Patel, Davis Rempe, Rulin Shao, Esin Tureci, Anya Tsvetkov, Jiachen T. Wang, Sheng-Yu Wang, Tingwu Wang, Zian Wang, Hongyu Wen, Jon Williams, Mengzhou Xia, Donglai Xiang, Yilun Xu, William Yang, and Haotian Zhang.

\newpage
\bibliographystyle{plainnat}
\bibliography{iclr2026_conference}

\appendix
\newpage
\clearpage

\section{Notation}
\label{app:notation}

\begin{table}[!ht]
\centering
\captionsetup{justification=centering}
\caption{Glossary and notation.}
\label{tab:notation}
\renewcommand{\arraystretch}{1.3}
\small
\begin{tabular}{l l}
\toprule
\textbf{Symbol} & \textbf{Description} \\
\midrule
\multicolumn{2}{l}{\textit{Acronyms and Basic Notation}} \\
VAE & Variational Autoencoder \\
DiT & Diffusion Transformer backbone \\
$\identity$ & Identity matrix \\
\midrule
\multicolumn{2}{l}{\textit{Video Generation}} \\
$\videogenerator(\videoclip \mid \conditioning)$ & Conditional video generator with parameters $\modelparams$ \\
$\videoclip \in \mathbb{R}^{\numFrames \times \videoHeight \times \videoWidth \times 3}$ & Video clip with frames $\numFrames$, height $\videoHeight$, width $\videoWidth$ \\
$\conditioning$ & Conditioning signal such as text or multimodal metadata \\
$\modelparams$ & Trainable model parameters \\
$\frameIndex \in \{1,\ldots,\numFrames\}$ & Frame index \\
$\videoHeightIndex \in \{1,\ldots,\videoHeight\}$, $\videoWidthIndex \in \{1,\ldots,\videoWidth\}$ & Spatial indices for height and width respectively\\
$\latentHeightIndex$, $\latentWidthIndex$ & Latent grid indices \\
\midrule
\multicolumn{2}{l}{\textit{Datasets}} \\
$\dataset = \{(\videoclip_{\datasetIndex}, \conditioning_{\datasetIndex})\}_{\datasetIndex=1}^{\datasetSize}$ & Training corpus with size $\datasetSize$ and index $\datasetIndex$ \\
$\datasetft \subseteq \dataset$ & Fine-tuning dataset \\
$\dataSubset \subseteq \dataset$ & Selected influential subset \\
$\influenceIndex \in \{1,\ldots,\influenceSubsetSize\}$ & The selected subset size\\
$\numMotions$ & Number of query data \\
$\motionIndex \in \{1,\ldots,\numMotions\}$ & Query index \\
$\inputData$ & Generic input data pair \\
$\testdata$, $\trainndata$ & Test/query pair and training pair \\
$\queryVideoclip$, $\queryConditioning$ & Query video and its conditioning \\
\midrule
\multicolumn{2}{l}{\textit{Latent Space and Diffusion Components}} \\
$\vaeencoder$ & VAE encoder \\
$\latents = \vaeencoder(\videoclip) \in \mathbb{R}^{\numFrames \times (\videoHeight/\latentDownsampleFactor) \times (\videoWidth/\latentDownsampleFactor) \times \latentChannelSize}$ & Latent video with spatial factor $\latentDownsampleFactor$ and channels $\latentChannelSize$ \\
$\noisyLatents$ & Noisy latent variable used in diffusion or flow matching \\
$\noisevector \sim \mathcal{N}(0,\identity)$ & Gaussian noise \\
$\noisepredict(\noisyLatents,\conditioning,\timestep)$ & Predicted noise network in diffusion training \\
$\vectorfield(\noisyLatents,\conditioning,\timestep)$ & Time-indexed vector field in flow matching \\
$\timederivative$ & Time derivative of the latent trajectory \\
$\schedulerCoeffA, \schedulerCoeffB$ & Scheduler signal and noise scales at timestep $\timestep$ \\
$\trainingtarget$ & Target noise or velocity used for supervision \\
$\timestep \in \{1,\ldots,\timestepSize\}$ & Diffusion or flow-matching timestep, with total timesteps $\timestepSize$ \\
$\timestepfixed$, $\fixednoise$ & Fixed timestep and shared noise draw used for low-variance gradients \\
\bottomrule
\end{tabular}
\end{table}

\begin{table}[!t]
\centering
\captionsetup{justification=centering}
\caption{Glossary and notation (continued).}
\label{tab:notation-attribution}
\renewcommand{\arraystretch}{1.3}
\small
\begin{tabular}{l l}
\toprule
\textbf{Symbol} & \textbf{Description} \\
\midrule
\multicolumn{2}{l}{\textit{Attribution and Influence}} \\
$\influenceFunc(\trainVideoData,\queryVideoclip;\modelparams)$ & Influence score between a train clip and a query clip \\
$\InfluenceMot(\trainVideoData,\queryVideoclip;\modelparams)$ & Motion-aware influence score \\
$\TopK(\cdot)$ & Top-$\influenceSubsetSize$ operator for selecting highest scores \\
$\MajVote(\cdot)$ & Majority-vote aggregation across queries \\
$\influenceThreshold$ & Percentile cutoff for voting \\
$\spearmancorr$ & Spearman correlation coefficient \\
\midrule
\multicolumn{2}{l}{\textit{Loss Functions}} \\
$\loss$ & Generic loss \\
$\lossdiff(\modelparams;\videoclip,\conditioning), \lossflow(\modelparams;\videoclip,\conditioning)$ & Diffusion and flow-matching objective \\
$\lossmotionweighted(\modelparams;\videoclip,\conditioning)$ & Motion-weighted objective used for attribution \\
$\perLocError$ & Per-location squared error in latent space \\
$\gradient$, $\gradientProjected$ & Gradient and its projected version \\
$\gradientMotion$, $\gradientMotionProjected$ & Motion-weighted gradient and its projection \\
$\hessianmatrix$ & Hessian with respect to $\modelparams$ \\
\midrule
\multicolumn{2}{l}{\textit{Motion Representations}} \\
$\alltracker(\videoclip) = \motioninfo$ & AllTracker motion extraction \\
$\motioninfo \in \motionspace$ & Motion tensor containing flow, visibility, and confidence \\
$\displacementvec(\videoHeightIndex,\videoWidthIndex)$ & Displacement vector at frame $\frameIndex$ and location $(\videoHeightIndex,\videoWidthIndex)$ \\
$\motionmagnitude(\videoHeightIndex,\videoWidthIndex)$ & Motion magnitude at a location, computed from the displacement \\
$\motionweights(\frameIndex,\videoHeightIndex,\videoWidthIndex) \in [0,1]$ & Normalized motion weights used to mask per-location losses \\
$\numericalBias$ & Small numerical bias for stability (e.g., $10^{-6}$) \\
\midrule
\multicolumn{2}{l}{\textit{Projections and Computational Details}} \\
$\gradientdim$, $\projecteddim$ & Full and projected gradient dimensions \\
$\projectionMatrix \in \mathbb{R}^{\projecteddim \times \gradientdim}$ & Projection matrix used for Fastfood-style JL projection \\
$\fastfoodNorm$ & Variance normalization constant for projection \\
$\timestepset$ & Set of sampled $(\timestep,\noisevector)$ pairs for gradient estimation \\
$\cost$ & Unit compute cost used in complexity accounting \\
\bottomrule
\end{tabular}
\end{table}

\clearpage
\vspace{-0.02\textheight}
\section{Extended Related Work}\label{sec:app:related}
\subsection{Data Attribution}
Understanding how individual training examples shape model behavior has been a long-standing goal. 
Modern data attribution methods fall into two main groups~\citep{hammoudeh2024training}: retraining-based methods (e.g., leave-one-out~\citep{cook1982residuals, jia2021scalability}, downsampling (also known as subsampling or counterfactual influence)~\citep{feldman2020neural}, Shapley-value~\citep{wang2024data, wang2024rethinking}) and gradient-based methods (influence-function family, including Influence Functions~\citep{koh2017understanding, lorraine2024jacnet}, TracIn~\citep{pruthi2020estimating}, and TRAK~\citep{park2023trak}).
Influence functions provide a principled framework by approximating the effect of removing a training point.
TracIn \citep{pruthi2020estimating} and TRAK \citep{park2023trak} make attribution feasible at scale. While effective for classification, these assume a mapping between training gradients and predictions, which becomes more complex in generative models.

Data attribution traces how individual training examples (or subsets) influence a model’s predictions or behavior. Formally, it assigns an attribution score to each training sample, estimating the extent to which that sample contributes (positively or negatively) to the model’s output on a given test query or behavior.
Influence data attribution is an example of nested optimization~\citep{vicol2022implicit, lorraine2024scalable} with other examples including differentiable games~\citep{balduzzi2018mechanics, lorraine2021lyapunov, lorraine2022complex}, hyperparameter optimization~\citep{raghu2021meta, lorraine2018stochastic, mehta2024improving, lorraine2020optimizing}, and variance-reduced gradient estimation for diffusion-based teacher--student distillation~\citep{bettencourt2026carv}.
Before diffusion models, attribution methods were applied to supervised learning tasks such as classification and regression, where influence functions \citep{koh2017understanding, lorraine2022task} and scalable approximations such as TracIn \citep{pruthi2020estimating}, TRAK \citep{park2023trak}, and TDA~\citep{bae2024training} quantified the impact of training examples on downstream predictions.
Recent work adapted data attribution to diffusion models~\citep{georgiev2023journey, zheng2023intriguing, wang2025fast, wang2024data, lin2024diffusion,brokman2024montrage, kwon2023datainf}, where iterative denoising introduces timestep-dependent bias. \citet{mlodozeniec2024influence} propose scalable approximations, while \citet{xie2024data} identify timestep-induced artifacts and normalization schemes. Concept-TRAK \citep{park2025concept} extends attribution to concepts by reweighting gradients with concept-specific rewards, enabling attribution to semantic factors. 
\citet{wang2023evaluating} instead design a customization-based benchmark for text-to-image models, where models are fine-tuned on exemplar images with novel tokens and attribution is evaluated by whether it can recover the responsible exemplars.
However, these are limited to image diffusion, which captures static appearance but not temporal dynamics.

\subsection{Motion in Video Generation}
Video diffusion extends image generation to time, requiring coherent motion across frames \citep{ho2022video, blattmann2023stable, peebles2023scalable, wan2025, agarwal2025cosmos}. A large body of work builds temporal structure via attention layers \citep{wu2023tune}, control signals \citep{chen2023control, zhang2023controllable}, feature correspondences \citep{geyer2023tokenflow, bao2023latentwarp, wang2024cove}, or consistency distillation \citep{wang2023videolcm, zhou2024upscale}. Recent work has highlighted the challenge of decoupling motion from appearance in video diffusion transformers, in which spatial and temporal information become entangled within the model's representations \citep{shi2025decouple}. However, understanding which training clips influence specific motion patterns in generated videos remains an open challenge.
 
In parallel, motion has long been studied using optical flow and correspondence, from classical formulations \citep{horn1981determining, lucas1981iterative} to modern approaches such as RAFT \citep{teed2020raft}, which improve accuracy and generalization. These priors are often repurposed during generation to guide dynamics, but they do not explain which training examples shaped a model’s motion behavior. Our work addresses both gaps by introducing a motion-aware data attribution framework specifically designed for video diffusion. We use motion-weighted gradients that disentangle temporal dynamics from static appearance, enabling us to trace generated motion patterns back to the most influential training clips.

\section{Additional Experiments}
\label{sec:add_experiments}
\subsection{Results on Additional Video Generation Models}

We further test our framework on additional video generation architectures beyond the \sbt{Wan} models shown in Sec.~\ref{sec:experiment} (Tab.~\ref{tab:vbench-wan}). We evaluate \methodname on \sbt{LTX-2B}~\cite{HaCohen2024LTXVideo}, which represents a different architectural design and training paradigm. The results in Tab.~\ref{tab:app-vbench-ltx} demonstrate that \methodname works effectively across different model architectures and scales.

\begin{table*}[t!]
    \centering
    \caption{\textbf{VBench Evaluation on LTX-2B.} Performance comparison on VBench~\citep{huang2024vbench} across different baselines (all values in $\%$, higher is better). All selection methods use $10\%$ of training data; MM: motion masking.
    }
    \label{tab:app-vbench-ltx}
    \small
    \renewcommand{\arraystretch}{1.12}
    \setlength{\tabcolsep}{5pt}
    \resizebox{\textwidth}{!}{%
    \begin{tabular}{@{}l*{6}{c}@{}}
    \toprule
    & \multicolumn{6}{c}{LTX-2B Video} \\
    \cmidrule(lr){2-7}
    Method & Subject Consist. & Background Consist. & Motion Smooth. & Dynamic Degree & Aesthetic Qual. & Imaging Qual. \\
    \midrule
Base        & 93.8 & 95.2 & 94.6 & 36.5 & 32.1 & 48.2 \\
Full FT    & 94.4 & \textbf{95.7} & 94.8 & 38.9 & 32.6 & \textbf{49.1} \\
Random     & 93.7 & 95.4 & 94.7 & 37.8 & \textbf{32.8} & 48.4 \\
Motion mag.& 94.1 & 95.3 & 95.2 & 39.6 & 32.4 & 47.9 \\
V-JEPA     & 94.2 & 95.5 & 95.1 & 40.2 & 31.9 & 47.5 \\
\addlinespace[2pt]
\cmidrule(lr){1-7}
Ours w/o MM& 94.3 & 95.6 & 95.3 & 41.8 & \textbf{32.8} & 48.7 \\
\rowcolor{ggreen!10}
\textbf{Ours (\methodname)}
           & \textbf{94.8} & 95.6 & \textbf{95.5} & \textbf{45.1} & 32.7 & 48.6 \\

    \bottomrule
    \end{tabular}
    \vspace{-20pt}
    }
\end{table*}

\section{Analysis}
\label{sec:analysis}

\begin{figure}[t]
    \centering
    \begin{subfigure}[b]{0.48\linewidth}
        \centering
        \begin{tikzpicture}
            \node (img1) {\includegraphics[width=\linewidth]{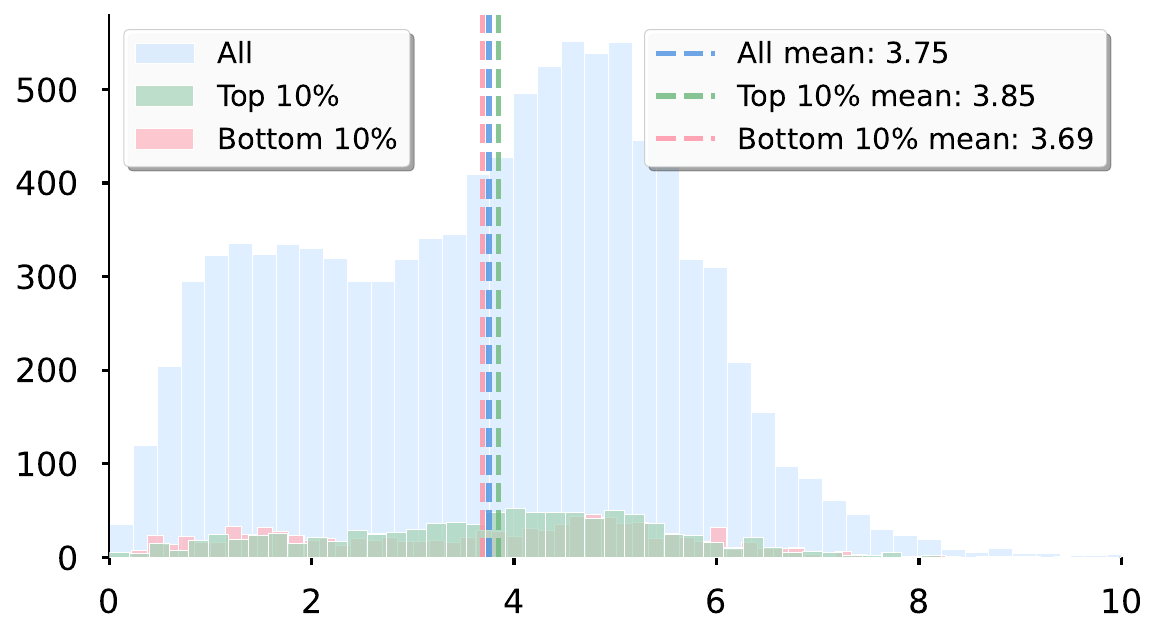}};
            \node[left=of img1, rotate=90, anchor=center, xshift=0.3cm, yshift=-0.9cm] {\footnotesize{Number of Samples}};
            \node[below=of img1, node distance=0cm, yshift=1.1cm] {\footnotesize{Mean Motion Magnitude}};
        \end{tikzpicture}
        \caption{Motion Distribution}
    \end{subfigure}
    \hfill
    \begin{subfigure}[b]{0.48\linewidth}
        \centering
        \begin{tikzpicture}
            \node (img2) {\includegraphics[width=\linewidth]{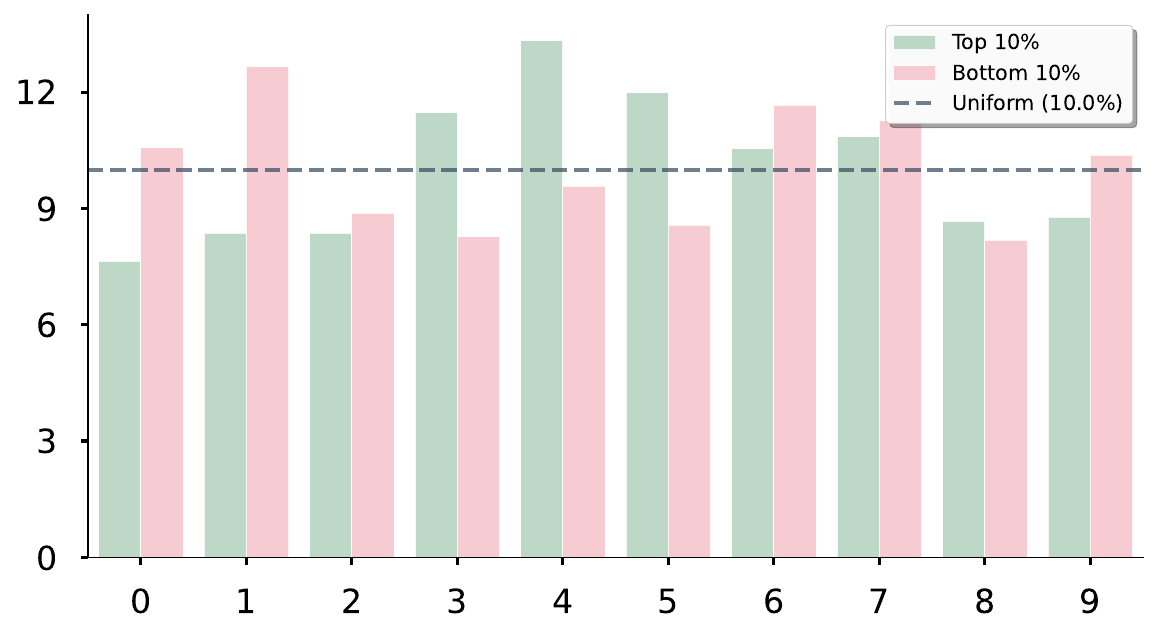}};
            \node[left=of img2, rotate=90, anchor=center, xshift=0.3cm, yshift=-0.9cm] {\footnotesize{\% of Selected Videos}};
            \node[below=of img2, node distance=0cm, yshift=1.1cm] {\footnotesize{Motion Bin}};
        \end{tikzpicture}
        \caption{Distribution Across Motion Bins}
    \end{subfigure}
    \vspace{-0.01\textheight}
    \caption{\textbf{\methodname is not simply selecting ``motion-rich" clips.} Our influence scores are computed via gradients, and training videos are considered influential only when they directly improve the model's ability to generate the target motion dynamics, not because they contain more motion overall.
    }
    \label{fig:motion_distribution}
\end{figure}
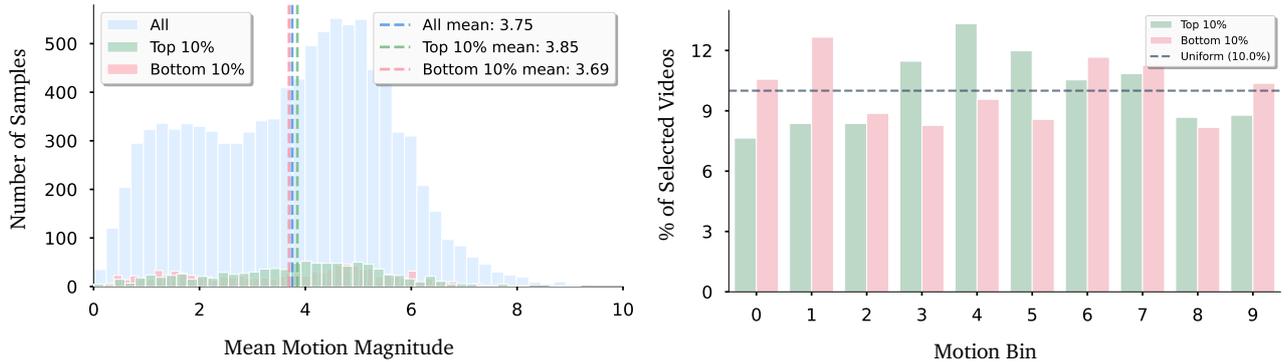

\subsection{Motion Distribution Analysis}

\textbf{\methodname is not simply selecting ``motion-rich" clips:} The key distinction is that our influence scores are computed via gradients, and training videos are considered influential only when they directly allow the model to lower the loss, improving the model's ability to generate the target motion dynamics, not because they contain more motion overall.

To empirically validate this, we further analyze the distribution of motion magnitudes in our selected data. We compute the mean motion magnitude for 10k videos in the \sbt{VIDGEN} dataset and compare the distributions of the top 10\% (highest influence scores) and the bottom 10\% (lowest influence scores).

As shown in Fig.~\ref{fig:motion_distribution}, the top 10\% selected videos have a mean motion magnitude of 3.85, which is only 4.3\% higher than the bottom 10\% (3.69), despite representing opposite extremes of influence scores. The analysis also shows that within the moderate-motion range (bins 3, 4, and 5), the top 10\% of positive-influence samples outnumber the bottom 10\% of negative-influence samples. Yet, both groups also appear in low-motion bins (0-2) and high-motion bins (6-9).

This distribution pattern shows that high-influence videos selected by \methodname span the entire motion spectrum, not just high-motion regions. Many high-motion videos receive low influence scores, while numerous influential videos exhibit modest or even low motion magnitude. These findings show that our motion attribution approach captures training influence, focusing on motion rather than simply acting as a motion-saliency filter.

\begin{figure}[t]
    \centering
    \begin{subfigure}[b]{0.48\linewidth}
        \centering
        \includegraphics[width=\linewidth]{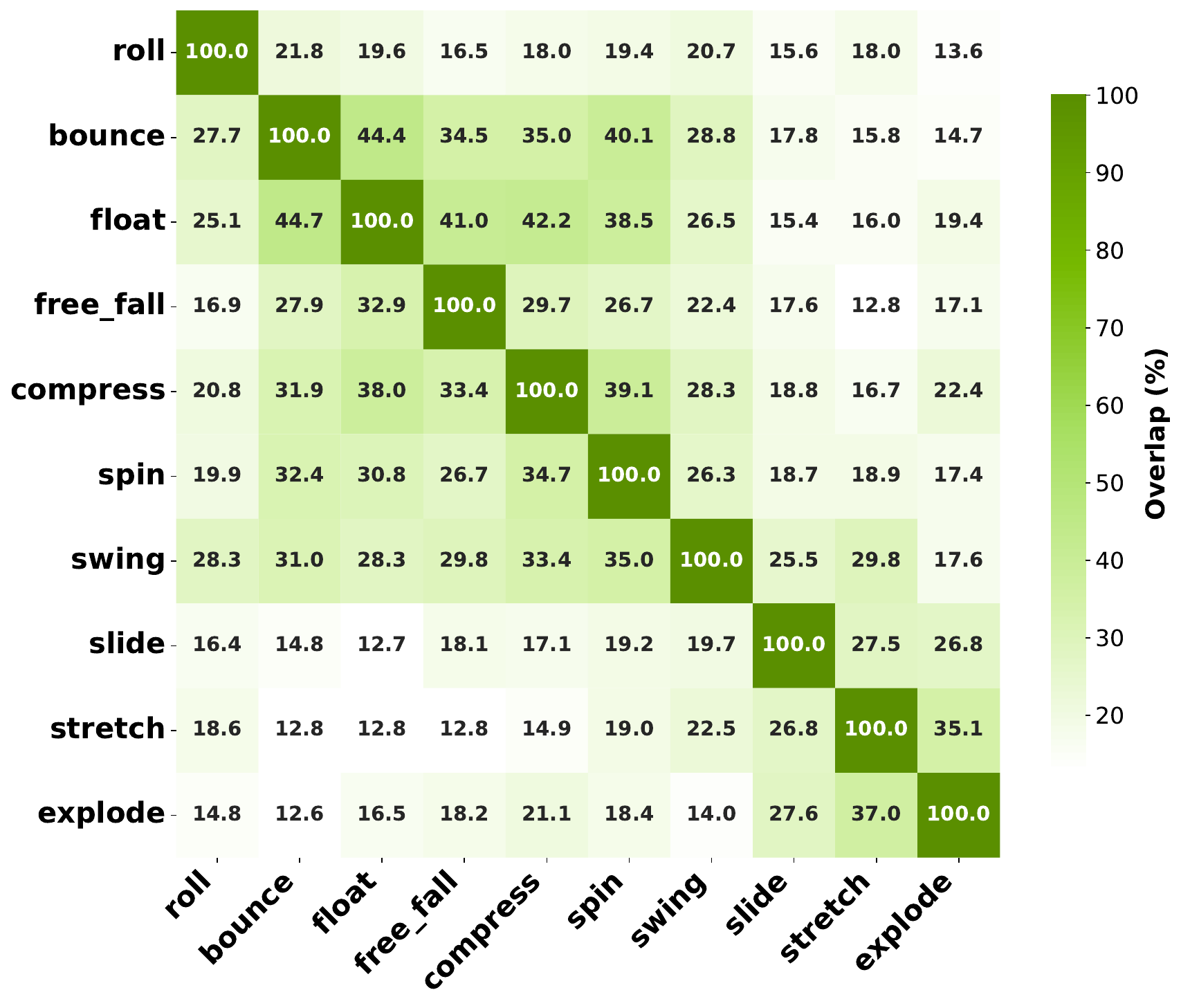}
        \caption{4DNEX Influence Heatmap}
        \label{fig:cross_motion_overlap_4dnex}
    \end{subfigure}
    \hfill
    \begin{subfigure}[b]{0.48\linewidth}
        \centering
        \includegraphics[width=\linewidth]{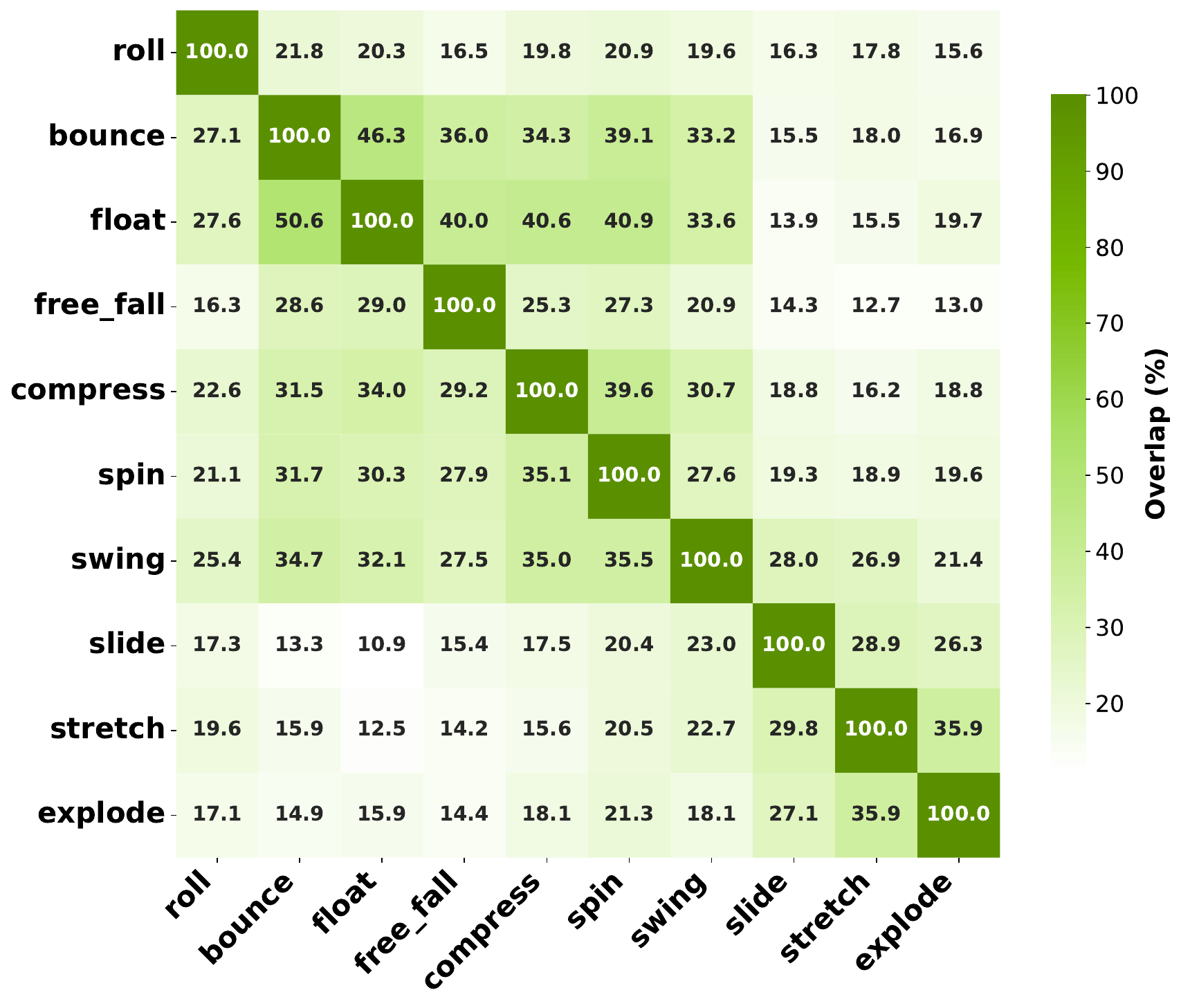}
        \caption{VIDGEN Influence Heatmap}
        \label{fig:cross_motion_overlap_vidgen}
    \end{subfigure}
    \caption{\textbf{Cross-motion influence overlap across datasets.} Heatmaps showing the percentage overlap of top-$100$ influential training samples across motion categories for (a) 4DNEX and (b) VIDGEN datasets. Each cell $(i,j)$ shows the percentage of motion category $i$'s influential data (aggregated from $5$ queries per category) that also appears in motion category $j$'s top-$100$ influential samples. The asymmetric nature of the matrices (e.g., bounce→float $\neq$ float→bounce) arises because different motion categories have different numbers of unique influential videos, leading to directional overlap percentages. Consistent high-overlap pairs (e.g., bounce-float: $44.4\%$/$46.3\%$) and low-overlap pairs (e.g., free fall-stretch: $12.8\%$/$12.7\%$) across datasets validate that these influence patterns reflect fundamental aspects of motion representation in video generation models.}
    \label{fig:cross_motion_overlap}
\end{figure}

\subsection{Cross-Motion Influence Patterns}

To analyze cross-motion influence patterns, we examine the percentage overlap of top-$100$ influential training data across different motion categories in both the 4DNEX and VIDGEN datasets. As described in \S\ref{sec:experiment}, we use $5$ query samples to identify the top-100 most influential training videos, aggregating results across queries. As shown in Fig.~\ref{fig:cross_motion_overlap}, both datasets exhibit remarkably similar patterns, with mean overlaps of $24.0\%$ and $24.3\%$, respectively, indicating a moderate degree of sharing of influential data across motion categories.

Both datasets consistently identify the same high-overlap pairs: bounce-float ($44.4\%$/$46.3\%$), compress-float ($40.1\%$/$34.0\%$), and compress-spin ($36.9\%$/$39.6\%$), suggesting these motions share fundamental characteristics that the model learns from similar training examples. Conversely, low-overlap pairs such as free fall-stretch ($12.8\%$/$12.7\%$) and float-slide ($14.0\%$/$10.9\%$) indicate more specialized influential data for mechanically dissimilar motions. The influence matrices are asymmetric because the number of unique influential samples shared among the $5$ query samples varies by motion category.
The similar cross-motion influence patterns observed across both the 4DNEX and VIDGEN datasets demonstrate that these relationships are generalizable across different video datasets and reflect dynamic similarity.

\section{Additional Method Details}
\label{app:algorithm}

\textbf{Tracker-agnostic scope.}
We treat the motion estimator as a pluggable source of saliency rather than a training dependency. 
Given displacement magnitudes, we construct latent-space weights via bilinear mapping and normalization. 
Our implementation supports alternative estimators (such as dense optical flow or point tracking) with identical interfaces, enabling users to swap AllTracker without modifying the attribution code.

\textbf{Model-agnostic scope.}
Our attribution only requires per-example gradients under matched $(\timestepfixed,\fixednoise)$, and therefore applies to both diffusion and flow-matching objectives. 
The score reduces to a gradient inner product under a fixed preconditioner; the generator architecture affects gradient statistics but not the definition of influence. 
In practice, replacing the denoiser or velocity field leaves the weighting and aggregation unchanged.

\textbf{Algorithm Summary.}
For completeness, Algorithm~\ref{alg:motion_attribution} summarizes the full \methodname pipeline, detailing the computation of motion-weighted gradients, projection into low-dimensional space, and the subsequent influence-based ranking and selection of training clips.

\begin{algorithm}[t]
    \caption{\methodname: Motion-Aware Data Attribution Framework}
    \label{alg:motion_attribution}
    \begin{algorithmic}[1]
    \Require fine-tuning corpus $\datasetft = \{(\videoclip_n, \conditioning_n)\}_{n=1}^{N}$, query video $(\queryVideoclip, \queryConditioning)$, fixed $(\timestepfixed, \fixednoise)$, projection matrix $\projectionMatrix$
    \Ensure Motion-aware influence scores $\{\InfluenceMot(\videoclip_n, \queryVideoclip)\}_{n=1}^{N}$
    
    \For{$(\videoclip_n, \conditioning_n) \in \datasetft$}
        \State $\motioninfo_n \leftarrow \textsc{AllTracker}(\videoclip_n)$ \Comment{Extract per-pixel flow displacements $\displacementvec$ (Eq.~\ref{eq:displacement-vector})}
        \State Downsample and normalize to latent-space motion mask $\motionweights_n$ (Eqs.~\ref{eq:motion-weight}--\ref{eq:per-location-error})
        \State Evaluate motion-weighted loss $\lossmotionweighted(\modelparams; \videoclip_n, \conditioning_n)$ (Eq.~\ref{eq:motion-loss})
        \State Compute motion gradient $\gradientMotion(\modelparams, \videoclip_n, \timestepfixed, \fixednoise) = \nabla_{\modelparams} \lossmotionweighted(\modelparams; \videoclip_n, \conditioning_n, \timestepfixed, \fixednoise)$
        \State Normalize by frame length: $\gradientMotion \leftarrow \gradientMotion /\numFrames$
        \State Project motion gradient: $\gradientMotionProjected(\modelparams, \videoclip_n) := \frac{\projectionMatrix\gradientMotion(\modelparams, \videoclip_n, \timestepfixed, \fixednoise)}{\|\projectionMatrix\gradientMotion(\modelparams, \videoclip_n, \timestepfixed, \fixednoise)\|}$ (Eq.~\ref{eq:diffusion-influence-motion})
    \EndFor
    \State Compute query gradient: $\gradientMotionProjected(\modelparams, \queryVideoclip)$ using the same procedure for $(\queryVideoclip, \queryConditioning)$
    \For{$n = 1, \dots, N$}
        \State $\InfluenceMot(\videoclip_n, \queryVideoclip) = \gradientMotionProjected(\modelparams, \queryVideoclip)^{\top} \gradientMotionProjected(\modelparams, \videoclip_n)$ (Eq.~\ref{eq:diffusion-influence-motion})
    \EndFor
    \State Rank all training clips by $\InfluenceMot(\videoclip_n, \queryVideoclip)$ and select top-$\influenceSubsetSize$ influential samples using majority vote aggregation (Eq.~\ref{eq:majority-vote}):
    \[
    \dataSubset = \dataSubset_{\textnormal{vote}}(\influenceSubsetSize) = \bigl\{\videoclip_n | \videoclip_n \text{ in top-}\influenceSubsetSize \text{ by } \MajVote \bigr\}
    \]
    \State \Return $\dataSubset$
    \end{algorithmic}
\end{algorithm}

\section{Additional Experiment Details}
\label{app:exp_details}

\subsection{Hyperparameter Settings}
\label{app:hyperparameter}

For reproducibility, we document the hyperparameters used throughout attribution, subset selection, and fine-tuning. Where values were not explicitly tuned, we adopted defaults from DiffSynth-Studio and the official Wan repository for Wan models and from Diffusion-Pipe for LTX models.

\textbf{Attribution.}
Motion-aware influence estimation is computed at a single fixed timestep, the midpoint of the denoising trajectory, which strongly correlates with multi-timestep averaging.
A shared Gaussian draw $\fixednoise \sim \mathcal{N}(0,\identity)$ is used across all training–query pairs to reduce stochastic variance. Gradients are projected from dimension $\gradientdim = \num{1418996800}$  to $\projecteddim = 512$ using a Fastfood Johnson–Lindenstrauss projection $\projectionMatrix$ selected via the search in Fig.\ref{fig:projection_correlation} to balance performance and storage. Motion weights $\motionweights$ are computed from AllTracker flow magnitudes $\motionmagnitude$, min–max normalized to $[0,1]$ with a small bias $\numericalBias = 10^{-6}$.
All computations use bfloat16 precision for memory efficiency.

\textbf{Subset Selection \& fine-tuning.}
For any number of query points, we select top-$10$\% data of the datasets. We finetune the backbone while freezing both the text encoder~\citep{raffel2020exploring} and the VAE. The input resolution is fixed to $480 \times 832$ pixels. We use a learning rate of $1 \times 10^{-5}$ and the AdamW optimizer~\citep{loshchilov2017decoupled} following the DiffSynth-Studio defaults. We train the models for $1$ epoch, repeating the dataset $50$ times.

\textbf{Evaluation.}
The test set consists of the same $10$ motion categories as the query set, but with different visual appearances. We provide the prompt samples below.  

\begin{example}[{\small Samples from Query Set}]
\small
We illustrate representative prompts from our query set used to generate query videos with Veo-3.
\begin{itemize}[leftmargin=*, itemsep=2pt, parsep=0pt]
    \item \textbf{compress}: ``A slice of sandwich bread flattened by a flat metal plate, steady camera, soft studio lighting, plain backdrop; emphasize air pockets collapsing.''
    \item \textbf{bounce}: ``A ping-pong ball bouncing on a white table, steady side camera, neutral light, seamless backdrop; emphasize consistent bounce height and timing.''
    \item \textbf{roll}: ``A spool of thread rolling from left to right, close-up static camera, bright studio light; highlight axle rotation and smooth travel.''
    \item \textbf{explode}: ``A single balloon bursting into fragments, captured in high-speed slow motion with a fixed camera, bright even lighting, seamless background; emphasize outward debris and air release.''
    \item \textbf{float}: ``A foam cube floating on the surface of water, static overhead camera, bright light, clean tank; emphasize buoyancy and slight rocking.''
\end{itemize}
\end{example}
\label{fig:query-data-samples}

\begin{example}[{\small Samples from Test Set}]
\small
We illustrate representative prompts from our test set that our fine-tuned models use to generate test videos.
\begin{itemize}[leftmargin=*, itemsep=2pt, parsep=0pt]
    \item \textbf{compress}: ``A rubber ball being compressed under a flat press, filmed with a stationary camera. Bright, shadow-free lighting and a clean background emphasize the deformation as it flattens.''
    \item \textbf{bounce}: ``A basketball bouncing vertically on a wooden court plank, unmoving camera, balanced indoor lighting, plain wall background; clearly show deformation at impact.''
    \item \textbf{roll}: ``A bike tire rolling freely on a stand, static side camera, indoor neutral light; show uniform rotation without wobble.''
    \item \textbf{explode}: ``A fragile glass ornament breaking apart mid-air, fixed camera, bright controlled lighting, plain backdrop; capture shards and reflections crisply.''
    \item \textbf{float}: ``A green leaf floating gently on perfectly still water in a transparent tank, fixed top-down camera, bright even lighting; emphasize surface tension ripples.''
\end{itemize}
\end{example}
\label{fig:test-data-samples}

\subsection{Details on Motion Query Data}
\label{app:motion_query_data}

\begin{figure}[t]
    \centering
    \includegraphics[width=\linewidth]{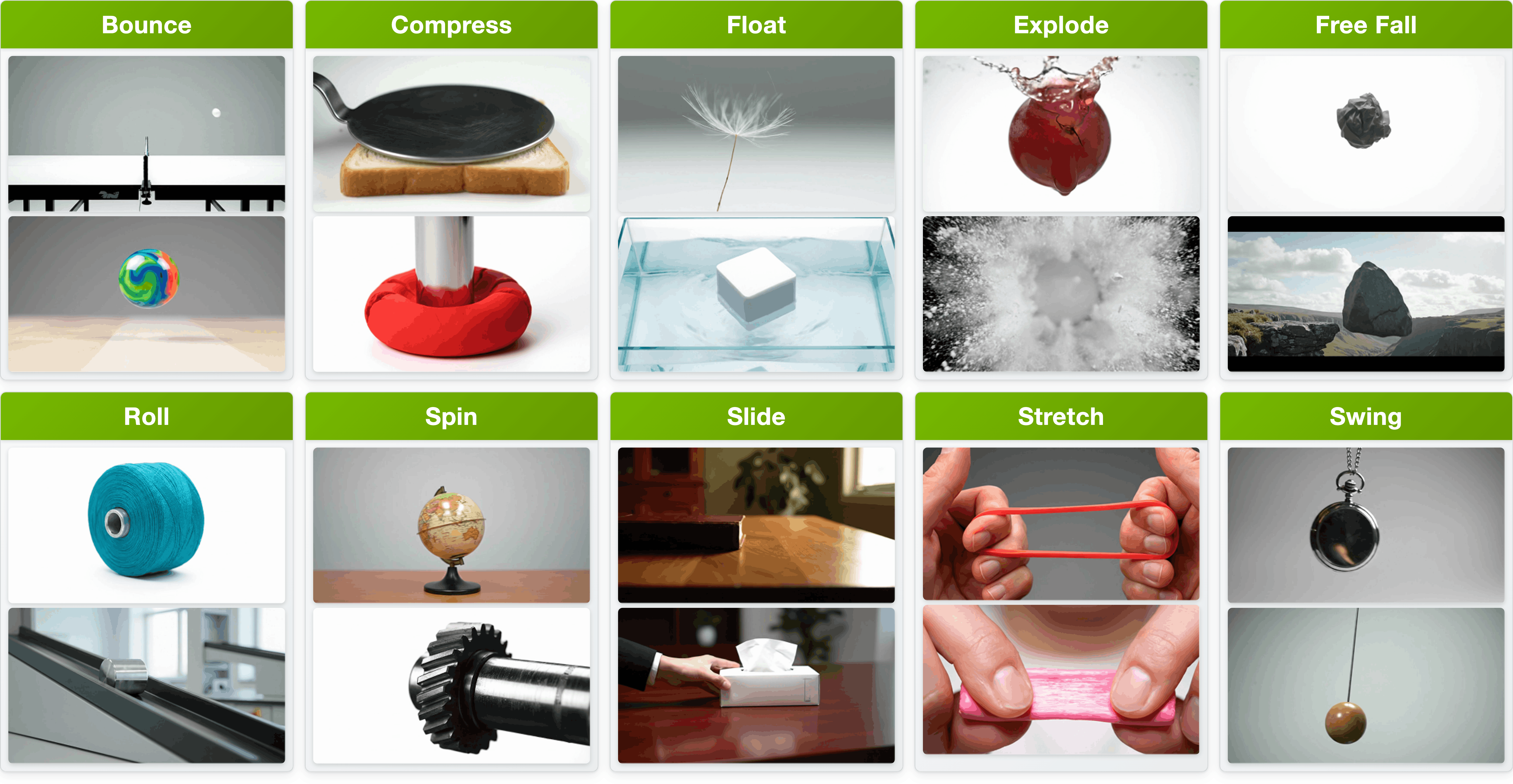}
    \vspace{-0.025\textheight}
    \caption{\textbf{Illustration of motion query set.} 
    We generate near-realistic video queries with Veo-3 across ten motion categories.
    Each category contains five query videos synthesized with controlled prompts and manually screened for clarity and physical plausibility.}
    \label{fig:veo3}
\end{figure}

A small, controlled set of query videos is constructed to isolate specific motion primitives while minimizing confounding factors (e.g., textured backgrounds, uncontrolled camera motion). Such clean and consistent clips are challenging to obtain from natural data sources. To address this, we synthesize the query set using Veo-3~\citep{veo3} and apply a strict post-generation screening for physical plausibility and generation realism. We target ten motion types: \emph{compress, bounce, roll, explode, float, free fall, slide, spin, stretch, swing}.
For each category, we retain $5$ query samples, yielding a total of $50$ queries. This scale provides adequate coverage of the motion taxonomy used in our evaluations while maintaining tractable attribution computation. We further provide a few examples of the generation prompts and the generated video query set in Fig.~\ref{fig:veo3}.

\textbf{Rationale for synthetic queries.}
The query set is not used as training data; instead, it specifies targets for attribution and for multi-query aggregation. Synthetic generation offers controllability that is difficult to achieve at scale with web videos. This design yields near-realistic yet standardized stimuli aligned with our goal of probing motion-specific influence.

\section{Discussion}
\label{app:discussion}

\subsection{Runtime}
\label{app:runtime}

\begin{table}[t!]
    \vspace{-0.01\textheight}
    \centering
    \caption{\textbf{Runtime Breakdown.} Detailed computational complexity and runtime for each component of our motion attribution framework on 10k training samples with \sbt{Wan2.1-T2V-1.3B} model.}
    \label{tab:runtime}
    \small
    \begin{tabular}{p{0.20\linewidth}|p{0.18\linewidth}|p{0.20\linewidth}|p{0.32\linewidth}}
    \toprule
    \textbf{Component} & \textbf{Complexity} & \textbf{Runtime} & \textbf{Notes} \\
    \midrule
    Gradient computation & $\mathcal{O}(\cost)$ per sample & Query: $\sim54$ seconds \newline Training: $\sim150$ hours  & $1$ A100 GPU; \newline Single forward+backward pass; \newline training is dominant cost but amortized over all queries; \newline
    embarrassingly parallel (with $64$ GPUs, $\sim2.3$ hours) \\ \midrule
    Projection & $\mathcal{O}(|\dataset| \cdot \projecteddim \log \projecteddim)$ & $\sim1.97$ seconds per sample & $\projecteddim = 512$ \\
    Influence computation & $\mathcal{O}(|\dataset| \cdot \projecteddim)$ & $\sim46$ milliseconds per query & $1$ query $\times$ 10k training samples \\
    Majority-vote aggregation & $\mathcal{O}(|\dataset| \cdot \motionIndex)$ & $\sim139$ milliseconds & $50$ queries $\times$ 10k samples \\
    \bottomrule
    \end{tabular}
\end{table}

All training runs are conducted on $4$-$8$ NVIDIA A100 GPUs.
We provide a detailed runtime breakdown from our experiments on 10k training samples with \sbt{Wan2.1-T2V-1.3B} model to address scalability concerns. The key insight is that the dominant cost of our pipeline is computing per-training-sample gradients, which is done once and can then be reused for all subsequent queries. Each training clip's gradient is projected into a compact $512$-dimensional vector, and adding a new query requires only (i) a single backward pass to obtain its own projected $512$-dimensional gradient and (ii) computing cosine similarity between the query vector and stored training vectors, which is exceptionally lightweight (on the order of seconds). Thus, the computational burden does not scale with the number of queries but only with the size of the training set.
As shown in Tab.~\ref{tab:runtime}, the dominant cost is the one-time training data gradient computation for 10k samples ($\sim150$ hours on 1 A100), which is amortized across all queries. Once computed, adding a new query requires only $\sim54$ seconds (gradient computation) + $46$ms (influence computation) = $\sim$54 seconds total. The training data gradient computation is embarrassingly parallel and can be reduced to $\sim\!\num{2.3}$ hours with $64$ GPUs.

\textbf{Runtime comparison with baselines.} We compare the computational cost of our method with the baseline approaches for processing 10k training samples on a single GPU (Table~\ref{tab:runtime-baseline}). While our method requires more upfront computation than the baseline approaches, this cost is amortized across all queries, and the computed gradients can be reused for multiple selection queries, making it practical for large-scale data curation scenarios.

\begin{table}[t!]
    \caption{\textbf{Runtime Comparison with Baselines.} Total computational time required for processing 10k training samples with \sbt{Wan2.1-T2V-1.3B} model on a single GPU across different data selection methods.}
    \centering
    \resizebox{0.9\linewidth}{!}{
    \begin{tabular}{lccccc}
    \toprule
    Method &
    Random &
    Motion Magnitude &
    Optical Flow &
    V-JEPA &
    Ours \\
    \midrule
    Total for 10k ($1$ GPU) &
    $<1$ second &
    $\sim5.5$ hours &
    $\sim5.7$ hours &
    $\sim3$ hours &
    $\sim150$ hours \\
    \bottomrule
    \end{tabular}
    }
    \label{tab:runtime-baseline}
\end{table}

\subsection{Limitations}
Gradient-based attribution is computationally expensive, requiring a high upfront cost for per-sample gradient computation (see \S\ref{app:runtime}), though this cost is amortized across queries.
Our analysis treats each video as a whole unit, thereby avoiding collapsing motion into frame-level appearance, but it risks overlooking the fact that only certain segments may carry motion-relevant information. Highly informative intervals can be diluted when averaged with static or redundant portions of the same clip. This suggests an open direction toward finer-grained attribution at the motion-segment or motion-event level, which could yield more precise insights into how different phases of a trajectory shape motion learning. Another limitation is that our motion masks may overemphasize camera-only motion; we detect this by spatial uniformity of $\motionweights$ and down-weight such clips, but a full disentanglement of ego and object motion remains future work.

Additionally, our framework does not explicitly account for classifier-free guidance (CFG), which is widely used in practice to steer video generation but introduces discrepancies between training-time attribution and inference-time dynamics. 
As a result, our influence estimates may not fully capture how guidance alters motion behavior.
In addition, while attribution-driven fine-tuning improves targeted motion quality, it may introduce trade-offs in the base model's capabilities.
This underscores the need for future work to balance targeted motion adaptation with the preservation of broader generative capabilities.

\subsection{Future Directions}

\textbf{Improving the attribution pipeline.}
Several directions can strengthen our motion attribution methodology.
\emph{Tracker-robust motion saliency}: replace or ensemble AllTracker with alternative estimators and use its confidence/visibility channels to weight masks.
\emph{Self-generated video queries}: use model-generated videos as queries to trace problematic motion patterns (e.g., unrealistic physics) back to training data, enabling iterative diagnostics and targeted motion improvement.

\textbf{Scaling data curation.}
\emph{Closed-loop data curation}: move from one-shot ranking to active selection: iteratively attribute, finetune, and re-attribute, or replace simple majority voting with learned query weights.
\emph{Sophisticated finetuning}: move to more sophisticated finetuning setups, such as multi-student distillation~\citep{song2024multi}.

\textbf{Extending to new domains.}
\emph{Other modalities}: extend our methodology to other modalities, including world models~\citep{zhu2025astra}, audio~\citep{evans2025stable,richter2025score}, or video+audio~\citep{wiedemer2025video}.

\textbf{Safety and governance.}
Use negative-influence filtering to suppress undesirable or unsafe dynamics, document curator choices, and audit motion behaviors that our framework exposes.

\section{Visualization}
\label{app:visualization}

\subsection{Motion Visualization}
\label{app:motion-visualization}

To provide intuition for the behavior of our motion-weighted loss, we visualize the motion magnitude as an overlay. We compute per-pixel motion magnitude using optical flow and apply motion-based weighting that preserves the appearance of dynamic regions while attenuating static backgrounds. This motion overlay directly illustrates the spatial weighting applied by our motion loss during training.

\begin{figure*}[t!]
    \centering
    \includegraphics[width=\linewidth]{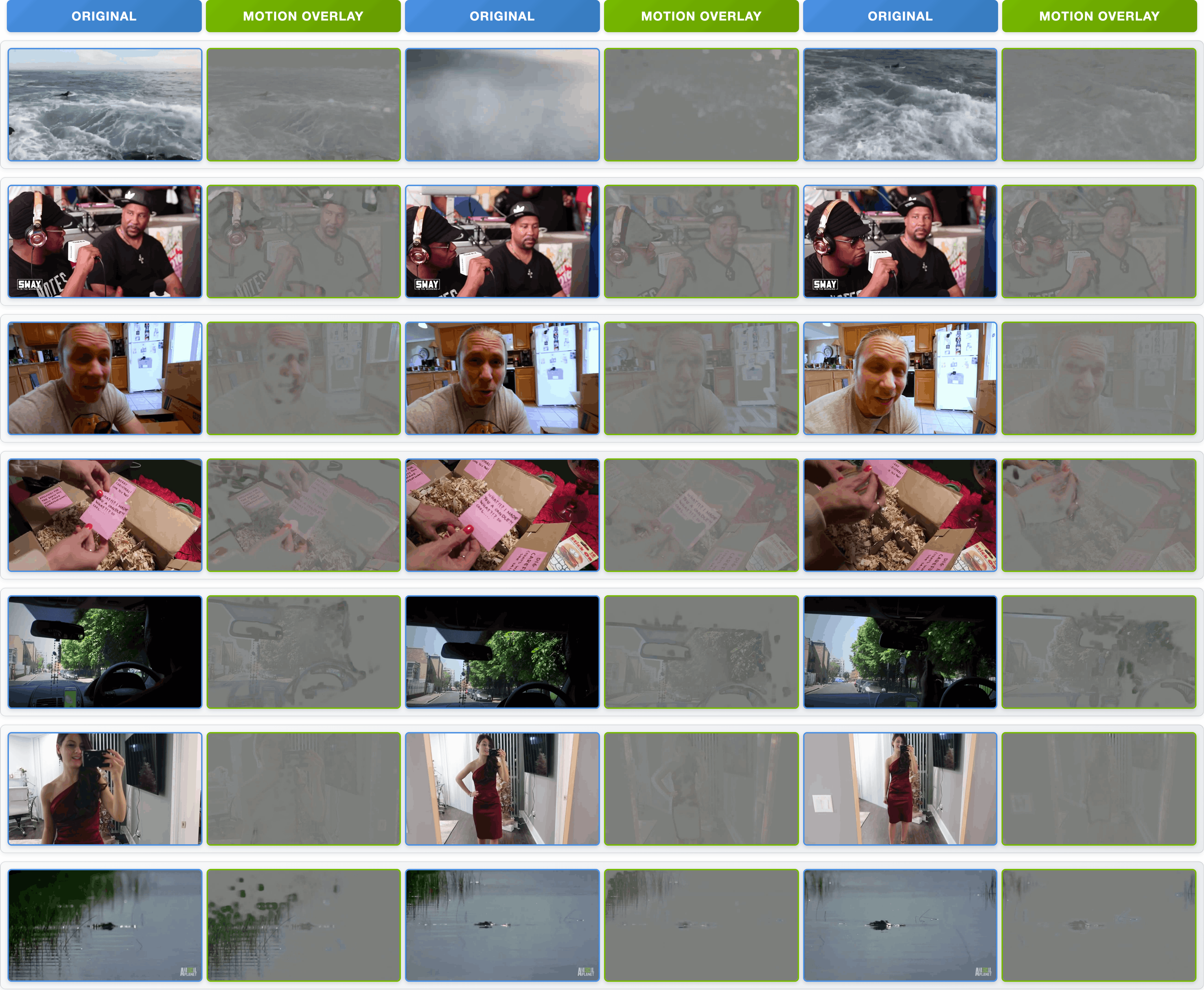}
    \caption{\textbf{Motion Overlay Visualization.} Comparison of original frames and motion overlays for seven video samples across three time points (early, middle, late). The motion overlay demonstrates the spatial weighting of our motion loss: dynamic regions remain visible, while static backgrounds are attenuated to neutral gray. \emph{Takeaway:} This provides heuristic intuition into what information our motion attribution focuses on: the information in grayer regions, which lack motion, is down-weighted by our method.}
    \label{fig:motion_examples}
\end{figure*}

Fig.~\ref{fig:motion_examples} presents representative frames from our dataset, comparing original frames with their corresponding motion overlays for seven distinct video samples. These visualizations show that the motion-weighted loss preferentially emphasizes dynamic content while down-weighting static scene elements.

\end{document}